\documentclass{article}
\counterwithout{figure}{section}

\usepackage[english]{babel}
\usepackage{amsmath}

\usepackage{authblk}
\usepackage[utf8]{inputenc}
\usepackage{etoolbox}
\usepackage{mathtools, nccmath}
\usepackage{lineno}

\usepackage[skip=5pt plus1pt, indent=0pt]{parskip}
\usepackage[letterpaper,top=2cm,bottom=2cm,left=3cm,right=3cm,marginparwidth=1.75cm]{geometry}
\usepackage{xcolor}

\usepackage{amsmath, amssymb}
\usepackage{graphicx}
\graphicspath{{Pics/}}
\usepackage{amsmath}
\DeclarePairedDelimiter{\nint}\lfloor\rceil
\usepackage[colorlinks=true, allcolors=blue]{hyperref}
\usepackage{xcolor, soul}
\sethlcolor{yellow}
\newcommand {\cA} {\mathcal{A}}
\newcommand {\cC} {\mathcal{C}}

\newcommand {\cS} {\mathcal{S}}
\newcommand {\bR} {\mathbf{R}}
\newcommand {\Rwrong} {R_{\rm{wrong}}}
\newcommand {\Rcorrect} {R_{\rm{correct}}}
\newcommand {\Rwait} {R_{\rm{wait}}}

\usepackage[labelsep=period]{caption}
\usepackage[labelfont=bf]{caption}
\makeatletter
\patchcmd{\affil}{\centering}{\raggedright\small}{}{}
\makeatother

\usepackage[backend=biber,style=numeric, sorting=none]{biblatex}
\bibliography{RLSS}

\title{Sequential sampling without comparison to boundary through model-free reinforcement learning}%

\newcommand{\correspondingauthor}{\textsuperscript{*}}

\author[1,2]{Jamal Esmaily\correspondingauthor}
\author[3,4,5]{Rani Moran}
\author[6,7]{Yasser Roudi}
\author[1]{Bahador Bahrami\correspondingauthor}

\affil[1]{Department of General Psychology and Education, Ludwig Maximilians University Munich, Munich, Germany}
\affil[2]{Graduate School of Systemic Neurosciences, Ludwig Maximilians University Munich, Munich, Germany}
\affil[3]{Max Planck University College London Centre for Computational Psychiatry and Ageing Research, University College London, Queen Square Institute of Neurology, London, UK}
\affil[4]{Wellcome Centre for Human Neuroimaging, University College London, Queen Square Institute of Neurology, London, UK}
\affil[5]{School of Biological and Behavioural Sciences, Queen Mary University of London, UK}
\affil[6]{Kavli Institute for Systems Neuroscience, Faculty of Medicine and Health Sciences, Norwegian University of Science and Technology, Trondheim, Norway}
\affil[7]{Department of Mathematics, King’s College London, London, United Kingdom}

\begin{document}
\date{}
\maketitle

\renewcommand{\thefootnote}{\fnsymbol{footnote}}
\footnotetext[1]{Corresponding authors: jimi.esmaily@gmail.com, bbahrami@gmail.com}

\begin{abstract}

Although evidence integration to the boundary model has successfully explained a wide range of behavioral and neural data in decision making under uncertainty, how animals learn and optimize the boundary remains unresolved. Here, we propose a model-free reinforcement learning algorithm for perceptual decisions under uncertainty that dispenses entirely with the concepts of decision boundary and evidence accumulation. Our model learns whether to commit to a decision given the available evidence or continue sampling information at a cost. We reproduced the canonical features of perceptual decision-making such as dependence of accuracy and reaction time on evidence strength, modulation of speed-accuracy trade-off by payoff regime, and many others. By unifying learning and decision making within the same framework, this model can account for unstable behavior during training as well as stabilized post-training behavior, opening the door to revisiting the extensive volumes of discarded training data in the decision science literature.


\end{abstract}
\section{Introduction}

Sequential sampling models have had great success in explaining the decision dynamics that govern the relationship between choice reaction time and accuracy under a variety of conditions spanning perceptual \cite{gold_neural_2007, britten_analysis_1992, roitman_response_2002}  , value-based  \cite{krajbich_multialternative_2011, mcginty_orbitofrontal_2016} and even moral decisions \cite{yu_how_2021}. 
The general principle of these models is that to make the best of the noisy, uncertainty-ridden information that an agent (e.g., rodent, monkey, human, etc) gets from its environment, one could accumulate the sequentially arriving noisy samples across time and compare the sum to a certain designated decision criterion. These models have been instrumental in interpreting the neurophysiological investigations of the mechanisms of decision making in humans \cite{kelly_internal_2013, ratcliff_quality_2009}, and non-human animals \cite{gold_neural_2007, brunton_rats_2013}. 

These models have often been applied to empirical data collected \textit{after} extensive training when performance has already stabilized at a predefined benchmark level and is unlikely to change with more practice.  However, a number of previous works examined the evolution of the drift-diffusion model (DDM) parameters in the course of learning \cite{ratcliff_aging_2006, masis_strategically_2023, uchida_speed_2003, shenhav_expected_2013, balci_acquisition_2011, liu_accounting_2012, dutilh_diffusion_2009}. Typically, these studies fit an instance of DDM to the empirically observed reaction time and choice data across the stages of learning. These results show that decision bounds decrease and/or drift rates increase during learning without explaining what mechanism could be implementing these changes. However, sequential sampling models are often kept agnostic about the process of how the agent comes to learn the decision criteria in the first place leaving open a question of how drift rates and/or thresholds are updated during learning.   

With adequate opportunity for practice, agents' behavior stabilizes in a given experimental context, for example under a specific payoff structure that relates choices to accuracy and rewards. Many studies have shown that agents are capable of changing their decision strategy, trading off speed and accuracy, when payoff structure is changed (see \cite{heitz_neural_2012} for a review; \cite{fleming_effects_2010, whiteley_implicit_2008};). Similar findings have documented decisional flexibility under other types of changes of context such as frequency of choice categories \cite{hanks_elapsed_2011} or the asymmetric physical effort needed for executing different choice options \cite{hagura_perceptual_2017}.  These behavioral adjustments to change of context have been attributed to the dynamic shifting of the decision boundary raising theoretical and empirical problems.

Theoretically, once the agent has learned a given decision boundary, it is not clear through what mechanism the boundary is subsequently adjusted in response to new contexts. A number of previous studies built on principles of model-based Reinforcement Learning (RL)  \cite{dayan_decision_2008, drugowitsch_cost_2012} to derive normative, “ideal-observer” solutions for what the boundary should be in order to maximize rewards or reward-rates, considering cognitive and opportunity costs associated with postponing a decision. These solutions assume that the agent has a "model" representing the statistical structure of the environment (e.g., the distribution of task difficulty). This model provides a transition structure that predicts the prospective amounts of evidence that would be accumulated if the decision were to be postponed. Based on such transition structures the agent can derive the optimal, reward-maximizing choice-threshold(s). This reward optimization problem was solved using the Markov decision process in one case  \cite{dayan_decision_2008} and Dynamic programming in another \cite{drugowitsch_cost_2012}. The key advantage of such model-based solutions is that they are highly flexible in that when the environment changes, they can quickly update their transition structure and readily recalculate the optimal choice threshold, without needing any elaborate experience with the new environment. However, a limitation of this model-based approach is that it relies on complex calculations that require a deep knowledge of the environment and the task at hand. It is unclear whether animals have access to such knowledge and can perform such demanding calculations. As a result, these approaches leave open the question of how to learn the boundary when such knowledge is not (yet) available.  

Empirically, the search for the neurobiological correlates of context-dependent shifting of decision boundaries has faced considerable difficulty. Numerous studies have identified specific neurons in the prefrontal and parietal cortex of various animals that show the hallmarks of the accumulation process during perceptual decision making under uncertainty. These neurons's activity rises during the period of stimulus observation. The rate of this rise is proportional to the strength of the sensory signal and reflects an accumulation of noisy evidence. This is at least the case in macaque area LIP where in trials with similar reaction times, these neurons reach a stereotyped firing rate shortly before action initiation; see \cite{gold_neural_2007, roitman_response_2002, brunton_rats_2013, hanks_neural_2014} for comprehensive review. The boundary shift hypothesis would predict that these stereotyped firing rates should covary with behavioral changes observed in speed-accuracy trade-off. This does not seem to be the case. Instead, Heitz et al. \cite{heitz_neural_2012} observed several heterogeneous phenomena (e.g., changes of baseline firing rate, sensory gain, and the duration of perceptual processing) in the activity of boundary neurons in the macaque monkey's frontal eye field during speed-accuracy trade-off. They noted that these observations were quite distinct from and in some cases even contradictory to the elegant and parsimonious predictions of a shift in the activity of the boundary neurons. Another study by Hank et al \cite{hanks_neural_2014} examined the neural activity in the macaque LIP with the hypothesis that this bound changes dynamically in response to different speed-accuracy trade-off conditions. They observed different results: the terminating threshold levels of neural activity were similar across all regimes even though the animal behavior adjusted dynamically to the different regimes. 

To address these problems, here we introduce a theory for learning how to make perceptual decisions under uncertainty based on model-free (temporal difference) RL principles. Our approach is simple and minimizes the required foreknowledge of the statistical structure of the environment compared to the model-based approaches discussed before. Our model also employs simpler calculations, but likely at the expense of flexibility \cite{sutton_reinforcement_2018}. Most importantly, our model dispenses with the concept of a decision boundary altogether. 

While standard RL models employ two actions corresponding to the two choice alternatives, our model adds to this an overtly simple innovation: besides the standard two actions, a Wait action permits the agent to \textit{stay undecided} and continue sampling the environment, albeit at a cost. We show that this minimal innovation creates a fundamentally new type of sequential sampling model that learns to make decisions under uncertainty and could dynamically change its strategy in response to changes of environment context. We demonstrate that this simple model reproduces the hallmark features of much more sophisticated evidence accumulation models.

\begin{figure}
\centering 
\includegraphics[width=1\textwidth]{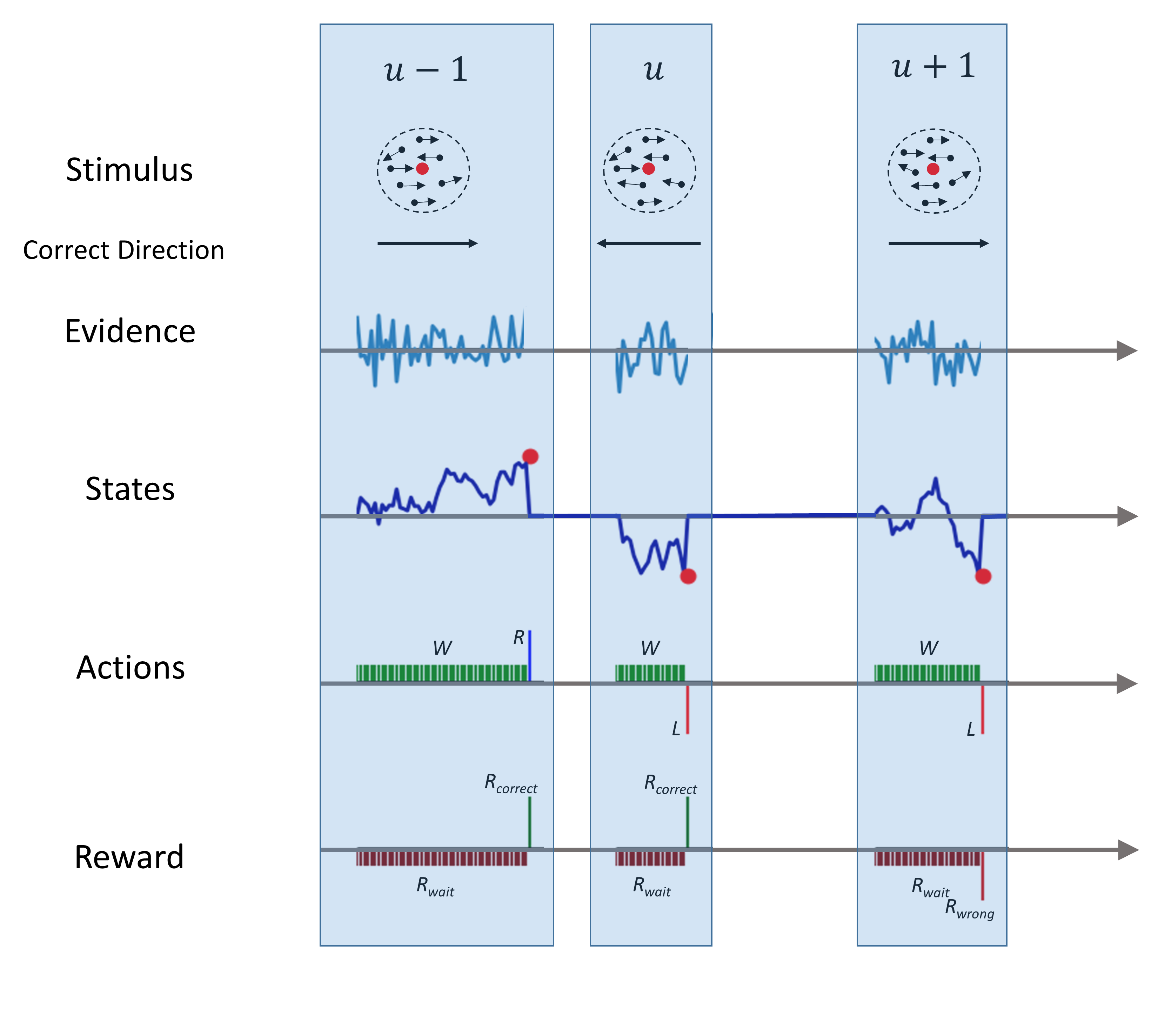}
\caption{{\bf Schematic illustration of the perceptual stimulus, trial structure, and model components.} The structure of the task and three consecutive trials ($u-1$, $u$, $u+1$) are illustrated with time progression taking place from left to right (horizontal arrow) both during the trials and from one trial to the next. The variable widths of the white gaps between trials depict the random duration of inter-trial intervals. We assume that no update happens during these periods. In each trial, a random dot motion stimulus moves towards the left or right. The evidence ($E_t$ in Eq.\ \eqref{eqn:state}) is sampled every time the agent chooses to wait (i.e., Wait action) and the state variable is updated by accumulating the evidence. This within-trial updating continues until the agent chooses one of the terminating actions (L, R) at which point the state and evidence variables are then set to zero and remain zero until the beginning of the presentation of the new motion stimulus in the next trial. The states at which these terminating actions are taken are the \textit{terminal state}, indicated by the red circle in the plots denoted by States. At each time point, the agent receives a reward based on the action that it has taken. Unlike sequential sampling models, no comparison to any threshold is explicitly formalized in the model and taken by the model.}
\label{fig:schema}
\end{figure}

\section{Results}
\subsection{The setup}
\label{sec:setup}
In Fig. \ref{fig:schema}, we show a schema of the model tailored to the two-alternative -- left vs right decisions-- random dot motion discrimination paradigm; see \cite{britten_analysis_1992, roitman_response_2002} for more details about the the task. In this model, time is defined from the agent's perspective. This description is agnostic to distinct notions of time (e.g., trial number, block number, trial onset, ...) that are meaningful only to the experimenter who is studying the agent. To simplify communication and avoid misunderstandings, in Fig.  \ref{fig:schema} and the model description that follows, we make this distinction explicit. We use separate notations to refer to the time – that we consider to be discrete – at the level of trial number and specific moments of time $within$ each trial, denoted by $u$ and $t$, respectively.

We consider a Q-learning agent trained over $U$ trials. In each trial $u$, a random dot motion stimulus is presented whose coherence level $c^u$ is sampled uniformly and independently from the set $\cC \subseteq [-1,1]$. Positive values of coherence indicate rightward motion and negative values indicate leftward motion. At any given time, the agent can choose an action from the set $\cA=\{\rm{Right, Left, Wait}\}$. If either actions Right or Left are chosen, then the agent receives a reward $\Rcorrect$ or $\Rwrong$ depending on whether the decision was correct or wrong, the current trial is terminated and a new trial begins. If the agent chooses the action Wait, it receives $\Rwait$ and the trial continues. Although for simplicity we refer to all these feedbacks as rewards, their values can indeed be negative and thus embody a cost. In what follows we assume $\Rwrong<0$, $\Rwait<0$ and $\Rcorrect>0$ unless stated otherwise; we often denote the set of reward values as $\bR \equiv [\Rcorrect,\Rwrong,\Rwait]$. Unless stated otherwise, $\bR \equiv [\Rcorrect = 20,\Rwrong = -50,\Rwait = -1]$

The agent is endowed with a set of states $\cS = \{-M,-M+\Delta \cdots, 0, \cdots, M-\Delta, M\}$. Here $\Delta$ indicates the resolution of the state of the model. At time $t$ in trial $u$, the state of the system is denoted by $s^u_t$ system. At the beginning of each trial the system starts at state $0$, that is $s^{u}_0=0$. As time progresses from $t$ to $t+\delta t$ within a trial, where $\delta t$ is the time step, that is while the agent chooses the Wait action, the state of the system is updated as
\begin{equation}  
\label{eqn:state}
s^u_{t+\delta t} = \nint{(s^{u}_{t} + E_{t})\delta t}_{\cS},
\end{equation}
where $\nint{x}_{\cS}$ indicated the closest element of $\cS$ to $x$, $E_{t} \sim \mathcal{N}(Kc^u,\,\sigma^{2})$ models noisy sensory evidence received at time $t$ and is taken to be a sample from a normal distribution with mean $Kc^u$ and variance $\sigma^2$. Unless otherwise stated, we use $\Delta = 1, \delta t = 1 ms$ (Therefore, the units of Reaction Time in our simulation are in milliseconds). 

Unless otherwise stated, in all simulations reported here, we have $K = 0.4$, and $\sigma = 1$. In principle, one can absorb $K$ into the range of stimuli coherence $\cC$, but to be consistent with previous studies where $K$ had to be fit to data, we employ the above notation. We also reiterate that the Left and Right actions lead to the termination of the existing trial, the start of a new trial, and are thus terminating actions. Before we proceed forward, we would like to note that although in Eq.\ \eqref{eqn:state}, we used the addition of the current state $s^u_t$ and sensory input $E_t$ for updating the state and for accumulation of evidence, this specific choice is not mandatory and the model can perform reasonably without accumulation too; see section \ref{sec:exterma}.

At any given time $t$, during the trial $u$, associated with every state $s\in \cS$ and every action $a \in \cA$, there is a Q-value denoted by $Q^{u}_t(s,a)$.  At the beginning of each trial, the Q-values for trial $u$ are copied from the end of trial $u-1$ with $Q^{0}_0(s,a) = 0$. The resulting Q-table is used to determine, in a given state $s$, which action, $a$, is selected at time $t$ in trial $u$. This is done via a softmax function yielding probabilistic action selection as follows
\begin{equation} 
\label{eqn:softmax}
p^{u}_t(a|s) = \frac{e^{\beta Q^{u}_t(s,a)}}{\sum_{a' \in \mathcal{A}} e^{\beta Q^{u}_t(s,a')}},
\end{equation}
where $\beta$ controls the degree of stochasticity in action selection and is set to $50$, unless otherwise stated. 
       
Once an action $a^{u}_t \in \{\rm{Right, Left, Wait}\}$ has been taken, the corresponding reward $R^{u}_t$ has been collected and the transition to the new state has occurred, the Q-table is updated as
\begin{equation} 
\label{eqn:Q-table}
Q^{u}_{t+1}(s^{u}_t,a^{u}_t) = Q^{u}_{t}(s^{u}_t,a^{u}_t)+\epsilon \left [R^{u}_t + \gamma  \max_{a\in \cA} (Q^{u}_{t}(s^{u}_{t+1},a)) -  Q^{u}_{t}(s^{u}_t,a^{u}_t) \right],
\end{equation}
where $\epsilon$ is the learning rate (set to 0.1 unless stated otherwise) and $\gamma$ is the discount factor of the temporal difference (TD) term. Unless otherwise stated, we use $\gamma =  0.9$ except when the trial is terminated and $\gamma = 0$. For the systematic study of the effect of parameters on the terminal state, see \ref{paramstudy}.
\subsection{Evolution of the Q-table during learning}
\label{sec:q_during}
In this section, we examine the evolution of the Q-table in the course of learning. A more detailed intuition about the dynamic of the model is provided in the Supplemental material (see \ref{sec:toy}).

\begin{figure}
\centering
\includegraphics[width=1\textwidth]{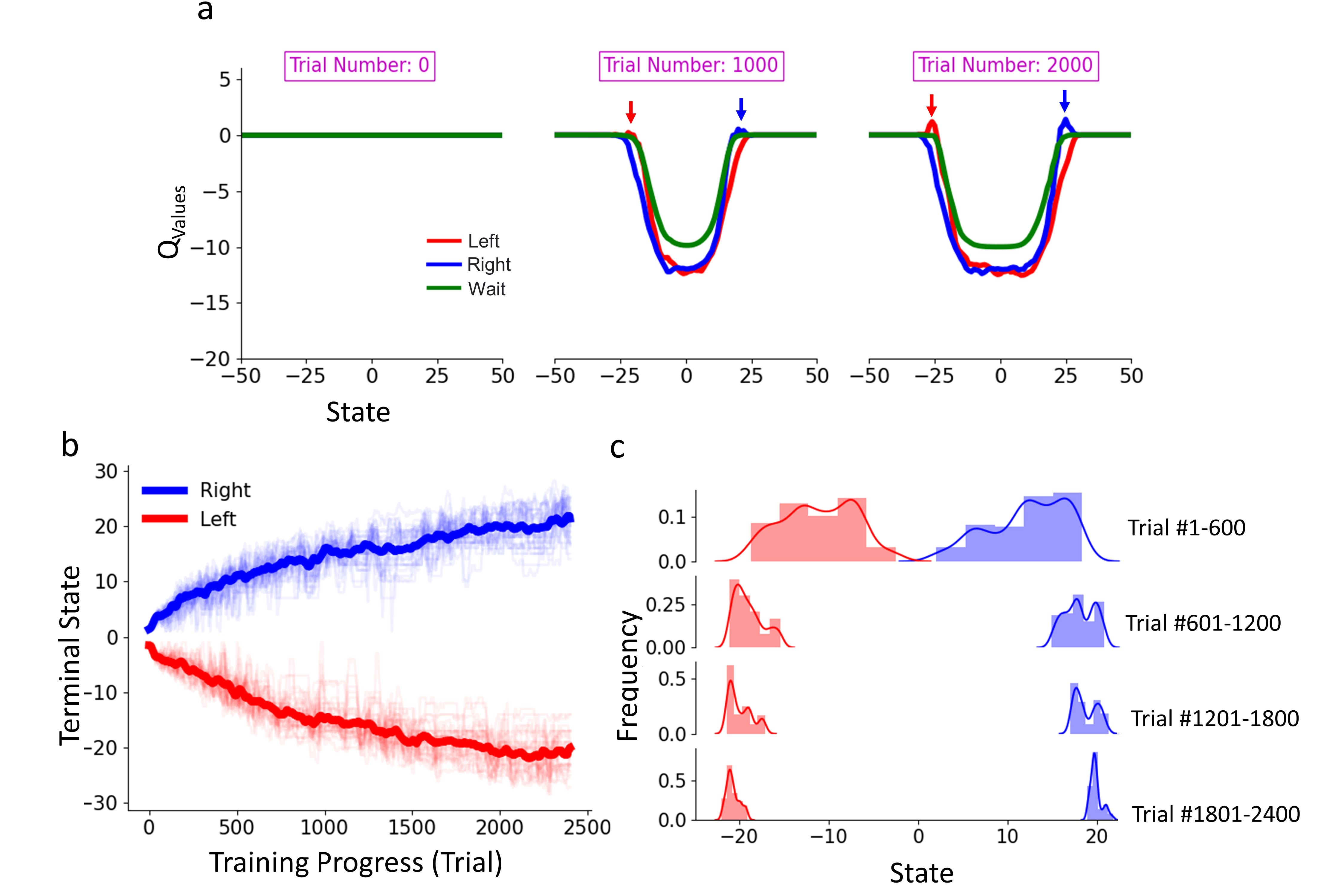}
\caption{{\bf Evolution of the Q-table.} (a) Snapshots of Q-values of each action at each state shown at the beginning of the learning (Trial number$=0$) where all Q-values are set to zero. The Q-values shown are averaged over 30 simulations with the same parameters. In each trial, $u$, the coherence level $c^{u}$ is chosen randomly and with equal probability from the set $\cC = [-51.2\%,-25.6\%, ...0,...25.6\%,51.2\% ]$. As training proceeds, the Q-values associated with the Wait action (green) in the states around zero stay higher but those for the terminating actions (Left, and Right) drop to lower values. The Q-value for each terminating action exceeds that of the Wait on the side corresponding to the correct choice (i.e., red on the left and blue on the right - see arrows). (b) Terminal states initially emerge near zero and then, with training, move away from it toward rightward (blue) and leftward (red). Each thin lines show the results for one of the 30 simulations, and the tick line represents the average over those simulations. (c) Histograms showing the fraction of times that a state had the largest Q-value (when averaged over 30 simulations) during 4 different periods of learning, each comprising $600$ trials. As training progresses, the histograms shift away from zero and become narrower in spread; the solid curves are fitted to the histograms. In the simulations reported in this figure, $U=2400$ trials were used.}
\label{fig:lr_q}
\end{figure}

In Fig. \ref{fig:lr_q}(a), we see three snapshots of the model Q-table at different stages of training. After 400 trials, there is an island of states around $0$, where the Q values of the Wait action (green) are the largest, and those of Right and Left are considerably more negative. As the learning proceeds (middle and right panels), this island expands. At the right boundary of this island, one can see a blue bump (c.f., blue arrows) indicating a number of states for which the corresponding Q-value for the Right action exceeds that for the other two.

Similarly, a red bump is visible on the left side, indicating those states for which the Left action is more likely to be chosen. How quickly in training these bumps appear depends on the learning rate, $\epsilon$; see Fig. \ref{fig:lr_sup}. By the end of the training, the Q-value for each of the terminating actions has exceeded that of the Wait action on its corresponding side, that is, red on the left and blue on the right. Note that, for the the peaks to arise stably by the end of the learning phase, the number of available states (i.e., $M$) should be large enough. With insufficient $M$, the Wait island (Green lines) expands to the whole range of available states, freezing the agent in a state of perpetual anticipation and paralysis.

We define the state in which a terminating action is chosen as the \textit{terminal state}. If $\beta = \infty$, the terminal states correspond to the peak of the bumps in the Q-table. For lower values of $\beta$, those peaks are merely more probable to be a terminal state. In Fig. \ref{fig:lr_q}(b)-(c), we see the evolution of the terminal states in the course of training. They start near zero and progressively move away. This trend is not monotonic, implying that the Q-learning algorithm searches for and ends up fluctuating around some Q-table that strikes a balance between the cost of waiting, the costs and benefits of wrong and correct decisions. 

\newpage

\begin{figure}
\centering
\includegraphics [width=0.9\textwidth]{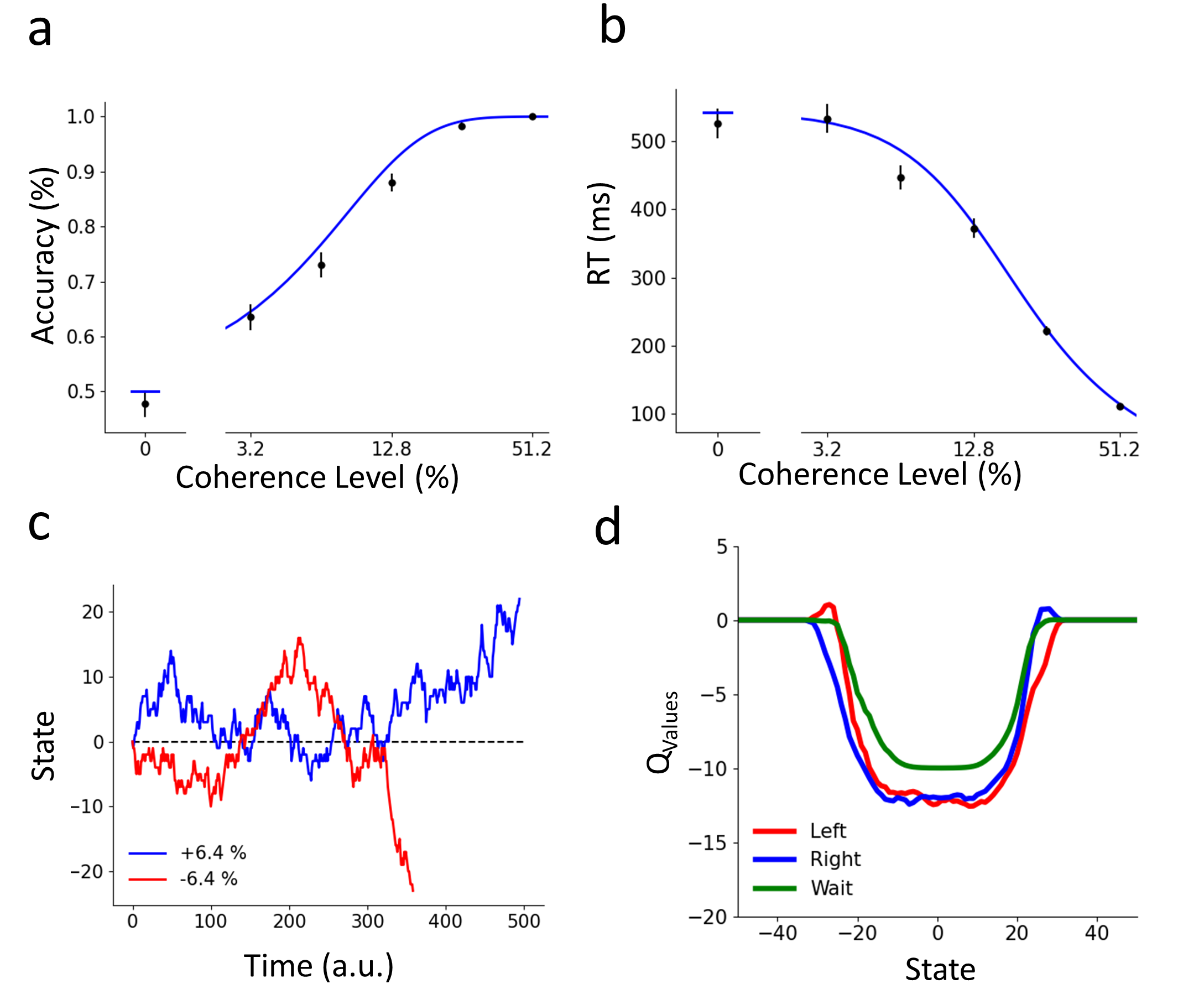}
\caption{{\bf Model performance after training.} The psychometric curve showing choice Accuracy (a), and the chronometric curve showing Reaction Time (RT) (b), both plotted as a function of the coherence level, $c$. Data from the simulations are denoted by black points and the lines in (a) and (b) show Eqs.\ \eqref{eqn:accan} and \eqref{eqn:rtan}. To plot these lines we fixed $B$ in Eq.\ \eqref{eqn:rtan} and \eqref{eqn:accan} to the average of the model's terminal state over $2400$ test trials during which the Q-table was left unchanged. The error bars show SEM over these trials. (c) Two examples of the states taken by the model as time progresses through the test trials where Q-table is fixed for two stimuli with opposing directions. (d) Q-values of the trained model for different actions in different states. The bumps appearing at states $\sim 20$ and $\sim -20$ indicate the location of the terminal states. All other parameters are the same as Fig. \ref{fig:lr_q}.
}
\label{fig:trained_model}
\end{figure}

\subsection{Model's behavior in perceptual decision making}
\label{sec:trained_model}
Having described some of the main features of the model above, here we showed that the trained RL model's behavior matches the hallmarks of perceptual decision making under uncertainty observed in empirical studies in humans, non-human primates, and rodents; c.f. \cite{gold_neural_2007,shadlen_decision_2013} for reviews of these empirical findings.

Fig. \ref{fig:trained_model}(a)-(b) shows the psychometric and chronometric functions that describe the relationship between decision accuracy and reaction time with motion coherence. As can be seen in this figure, increasing coherence increases accuracy and decreases reaction time, replicating extensive previous empirical findings. Perhaps more surprisingly, the simulation results (black symbols) are also in decent agreement with predictions obtained directly from the closed-form solutions to the bounded accumulation process \cite{ratcliff_theory_1978}:

\begin{subequations}
\begin{align}
\begin{split}
{\rm Accuracy}(c|B,K) =\frac{1} {1+ \exp[-2KcB]}
\label{eqn:accan}
\end{split}\\
\begin{split}
{\rm RT}(c|B,K) =\frac{B} {Kc}\ \tanh(KcB),
\label{eqn:rtan}
\end{split}
\end{align}
\label{eqn:sat}
\end{subequations}
where $K$ and $c$ were already introduced in Eq.\ \eqref{eqn:state}, and $B$ is the terminal state of the RL model at the end of learning, defined in the section \ref{sec:q_during}. Previous empirical studies that employ the sequential sampling models often interpret the behavior using the above two equations for which $K$ and $B$ are fitted to the data. In the language of sequential sampling,  \textit{K} and \textit{B} are known as the drift rate coefficient and the decision bound/threshold, respectively.

In addition to demonstrating the model’s summary behavior through psychometric and chronometric functions, we also studied the inner workings of the model within the course of each trial, as shown in Fig. \ref{fig:trained_model}(c). We see the state transitions (Eq. \eqref{eqn:state}) in two example trials. The blue trace shows a trial in which a weak rightward ($c=+6.4\%$) stimulus was presented to the model. The model took its time, collecting evidence and switching states for a fairly long number of time ($\approx 500$) steps. In comparison, the red trace shows another trial where a similarly weak but leftward stimulus was presented to the model. Here the model took a shorter time, arriving at the correct terminal state before the $400$ time steps. These two examples suggest that our RL model performs similarly to the sequential sampling models. This is particularly remarkable because at no point in the model description and training did we introduce any explicit boundary. This is where our model diverges from the ideal observer approaches  \cite{drugowitsch_cost_2012, dayan_decision_2008, skatova_extraversion_2013} that calculate the boundary based on their \textit{a priori} knowledge of the structure of the environment. 

In \ref{sec:replic}, we demonstrate the replication of a number of other, related empirical observations such as  the difference between error and correct reaction times (Fig. \ref{fig:err_corr}), post-error slowing (Fig. \ref{fig:PES}) and the impact of volatility on decisions accuracy and reaction times (Fig. \ref{fig:vol_replicate}). We encourage interested readers to utilize the model code and try out replicating other empirical findings in perceptual decision-making.


\section{Decision dynamics during learning}
\label{sec:training}
A key problem with previous RL models of perceptual decision making \cite{law_reinforcement_2009, kahnt_perceptual_2011} is that they did not produce any predictions about the development of reaction times during training. Our model, however, is naturally apt to address this problem. We, therefore, proceeded to examine how model reaction times and accuracy change in the course of learning. 

\begin{figure}
\centering
\includegraphics[width=.9\textwidth]{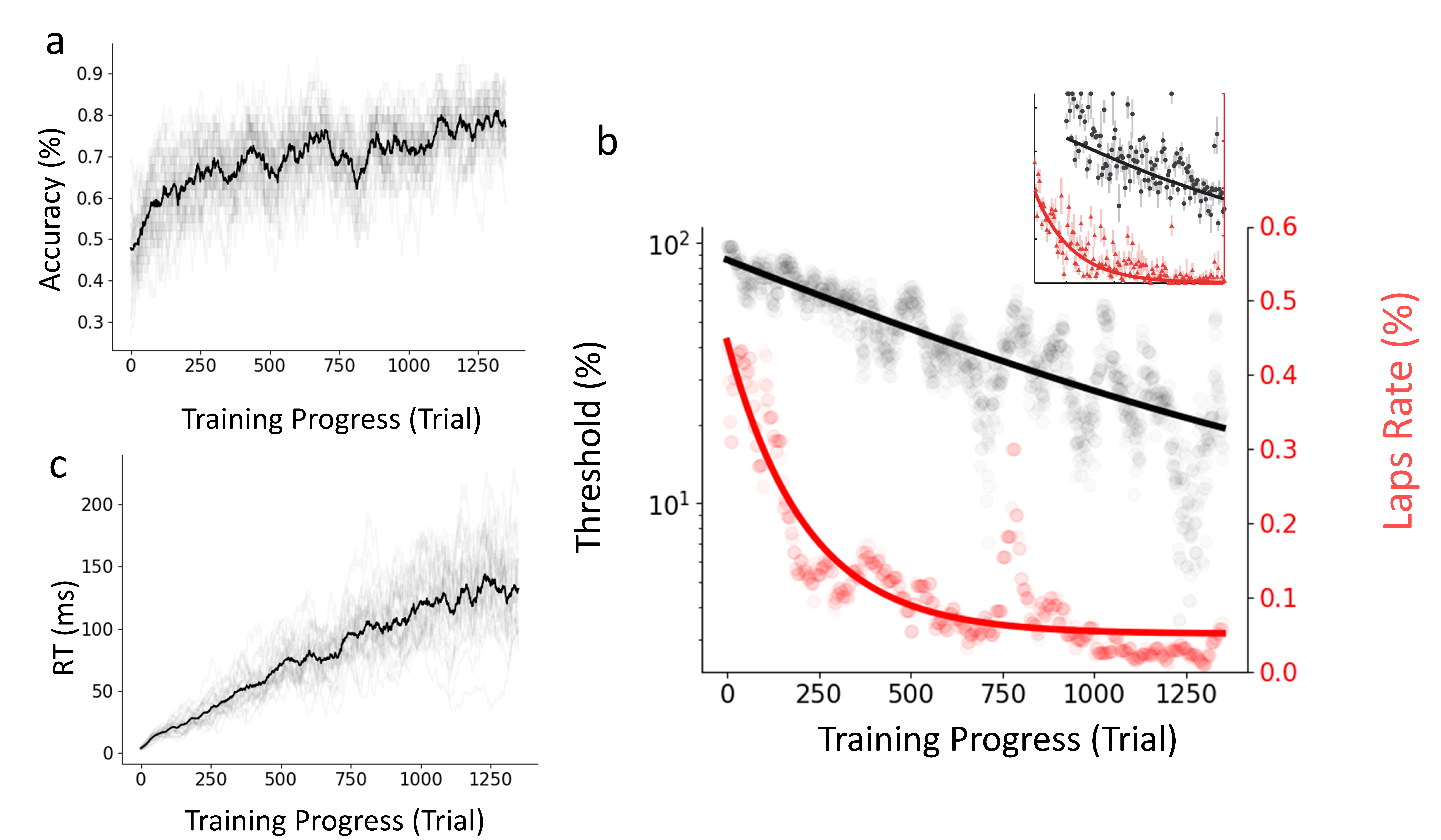}
\caption{{\bf Changes in decision Accuracy and RT during training}. (a) Accuracy increases as training progresses. Light grey curves show this for 30 individual simulations, with the same model parameters, smoothed through convolution with a unity array of size of 50. The solid black line shows the average over these simulations. (b) Changes in psychometric threshold (black) and lapse rate (red ) during training. The psychometric threshold is defined as the coherence level at which the model performs at 82$\%$ accuracy (e.g. $\alpha$ in Weibull CDF \cite{roitman_response_2002, law_neural_2008}) and the lapse rate is the error rate in trials with 100$\%$ coherence. Think curves are fits to the data points via similar functions used in \cite{law_neural_2008}. The inset shows empirical data from macaque monkeys \cite{law_neural_2008}. (c) Same as (a) but for the reaction times.}
\label{fig:BehavebyTime} 
\end{figure}


Fig. \ref{fig:BehavebyTime}(a) illustrates the changes in model decision accuracy with progress in learning. Consistent with numerous empirical observations, model accuracy starts around the chance level and progressively improves. We draw a direct comparison between the model behavior in Fig.  \ref{fig:BehavebyTime}(b) and empirical data reproduced from a previous study \cite{law_neural_2008} in the inset. Adopting that study's terminology, the threshold was defined as the coherence level at which the model performed at $82 \%$ accuracy, and the lapse rate was defined as the residual error rate at the highest level of coherence ($100 \%$). Our model’s choice behavior shows qualitative consistency with those empirical observations. 

In Fig. \ref{fig:BehavebyTime}(c) the RL model’s reaction times in the course of perceptual learning are plotted. Reaction times start fast and slow down with more training. This pattern of reaction times is indeed a direct consequence of the Q-learning algorithm: since the Q-table is initialized to zero, actions have similar values at the onset of learning. Therefore, in the early phases of learning, the terminating Right or Left actions are chosen with a high probability of  $\sim 66.6\%$ before any evidence is accumulated. The model learns to wait by committing frequent quick errors and decreasing the value of all three actions for the states around zero, albeit to different degrees. In other words, the consequence of starting the Q-table with such a blank slate is that the model would be barely exposed to the stimulus early in training. Consequently, early trials provide little opportunity for the model to learn the association between coherence and terminal actions. This profile of behavior, however, is different from several previous empirical observations where training usually starts with slower reaction times that progressively get faster \cite{ratcliff_aging_2006, masis_strategically_2023, uchida_speed_2003, shenhav_expected_2013, balci_acquisition_2011, liu_accounting_2012, dutilh_diffusion_2009}. Since this divergence is largely a consequence of the initial symmetry of the actions embodied in the initial blank slate Q-table, in \ref{sec:stlr}, we offer two alternative solutions to this issue based on the initialization of the Q-table differently. 

\begin{figure}[h!]
\centering
\includegraphics[width=.8\textwidth]{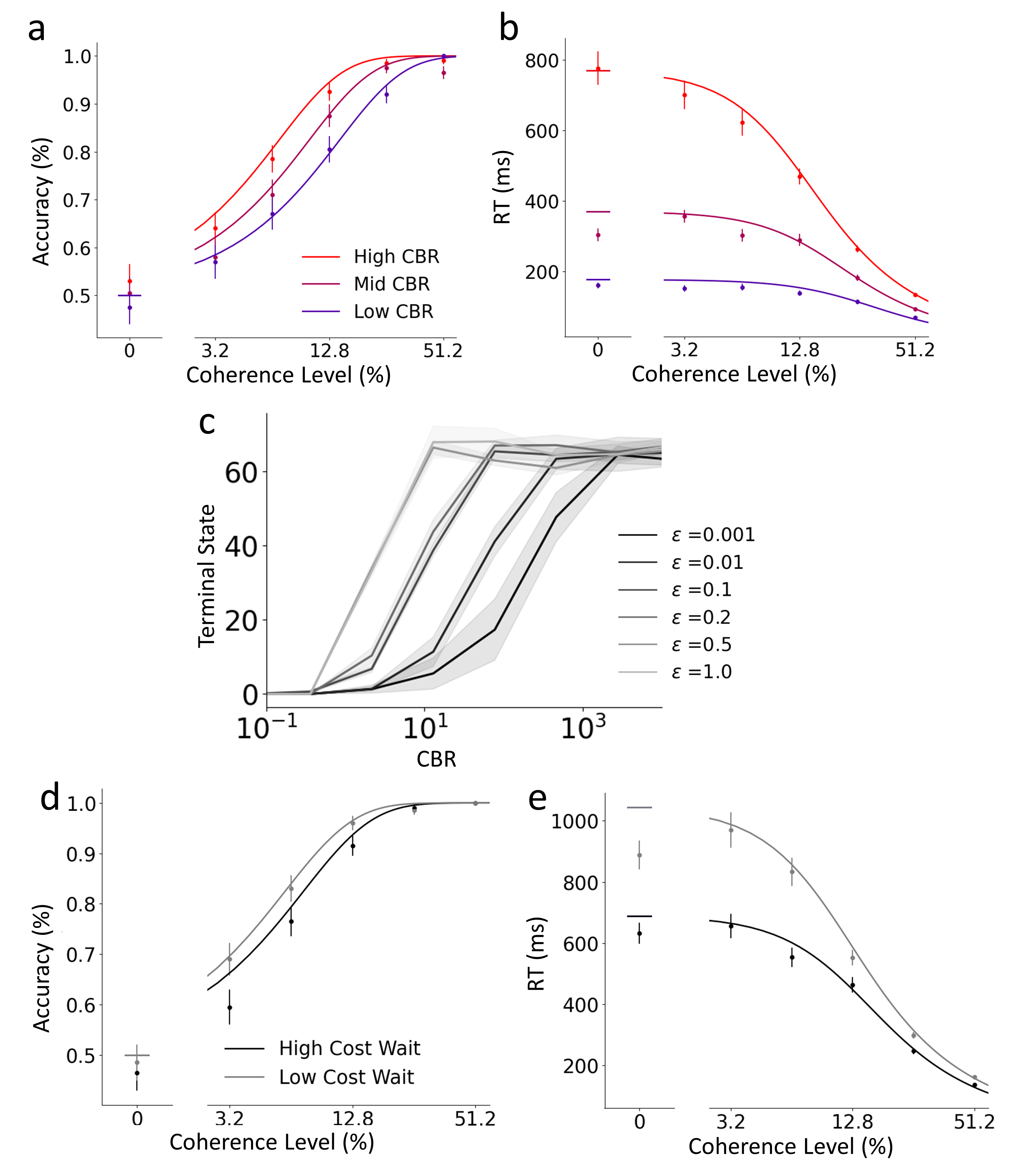}
\caption{{\bf Speed accuracy trade-off (SAT) of the trained model} (a) Choice Accuracy and (b) RT for different values of CBR: Large (4), Intermediate (3), and Small (2) indicated by different colors. (c) The terminal state for different values of CBR and different learning rates $\epsilon$. Increasing  CBR pushes the terminal state further away from zero, producing the dependence shown in (a)-(b). Curves are smoothed using a moving average filter and $ U = 900$. CBR values were changed from $0.01$ to $10^{5}$ in equal logarithmic steps, by fixing $\Rcorrect = 20$ and changing $\Rwrong$. Choice Accuracy (d) and RT (e) versus coherence level the cost of the Wait action is changed while CBR is kept constant. RT values are smaller and Accuracy is lower compared to the cost of the Wait action, as can be seen by comparing black and gray curves corresponding to $\Rwait = -2$ and$\Rwait = -1$, respectively. Error bars are SEMs across trials; The simulation involved 1200 trials.}
\label{fig:SpeedAccT}
\end{figure}

\section{The impact of payoff structure on Speed-Accuracy Trade-off}
\label{sec:SAT}
Having examined the dynamics of Speed-Accuracy Trade-off (SAT) during learning, we then proceeded to examine whether the model could flexibly trade-off speed and accuracy under various payoff conditions. This demonstration is critical for two reasons. First, previous reinforcement learning models have never been employed to explain the variations in choice reaction time in response to characteristics of the environment. Second, the extensive previous literature on speed-accuracy tradeoff 
\cite{heitz_speed-accuracy_2014, wald_sequential_1947, stone_models_1960}  in experiments involving human \cite{simen_rapid_2006, bogacz_humans_2010} and non-human \cite{heitz_neural_2012} primates provides a strong set of constraints to test our model with. 

Here, we focus on previous empirical works in humans demonstrating that increasing the cost of errors relative to the reward for correct choice prolongs reaction times and prioritizes accuracy; conversely, speed was prioritized when the reward for correct choice was increased \cite{simen_rapid_2006, simen_reward_2009, bogacz_humans_2010}.  These empirical observations were explained by changes of bound in sequential sampling models with a fixed drift rate. Several other studies that examined SAT in humans in a variety of decision tasks have also argued that SAT is best explained by changes in decision bound \cite{reddi_influence_2000, reddi_accuracy_2003}; c.f. Discussion).

To examine if our modeling framework could account for these empirical observations, we examined the following hypothesis: as long as the cost of waiting is kept low, increasing the cost of mistakes (vs the benefit of correct responses) should tip the balance towards accuracy. We tested our model under various payoff regimes, systematically altering the Cost-Benefit Ratio (CBR), defined as $|\Rwrong/\Rcorrect|$, over a wide range without imposing or assuming any decision threshold beforehand. 
The results in Fig. \ref{fig:SpeedAccT} show that when CBR is high (red curves in Fig. \ref{fig:SpeedAccT}(a)-(b)), the model reaction times are longer and accuracy is higher. In contrast, when CBR is low, decisions are faster and mistakes are more frequent (purple curves in Fig. \ref{fig:SpeedAccT}(a)-(b). These findings confirm our hypothesis. To further understand how these results arise from the RL dynamics, we investigated the relationship between CBR and the position of terminal states in the Q-table. Fig. \ref{fig:SpeedAccT}(c) shows a direct relation between the position of the terminal state and the CBR, depending on the learning rate $\epsilon$. For any given $\epsilon$, the position of the terminal state remains relatively constant for CBRs beyond a certain critical value. This value obviously depends on $M$ and also other parameters of the model, e.g. the total number of trials (see Fig. \ref{fig:opt_trial} for more details).

One caveat of examining SAT as a function of CBR is that CBR is independent of the cost of the Wait action. However, any plausible mechanistic explanation of SAT should factor in this cost \cite{drugowitsch_cost_2012}. To examine the impact of changing the cost of the wait (W) action on SAT balance, we tested the hypothesis that with equal cost of error and benefit of correct response, increasing the cost of waiting should prioritize speed. Fig. \ref{fig:SpeedAccT}(d)-(e) show that indeed, when waiting is more costly (black curve), reaction times decrease and accuracy is diminished. Reducing the cost of waiting (gray curve in Fig.  \ref{fig:SpeedAccT}(d)-(e)) reverts the trade-off in favor of accuracy. Together, the results in this section indicate that our model-free RL is able to \textit{learn} how long to wait before committing to a definitive choice in a way that balances the cost of evidence accumulation against the cost and benefit of choice outcomes.


\begin{figure}[h]
\centering
\includegraphics[width=.99\textwidth]{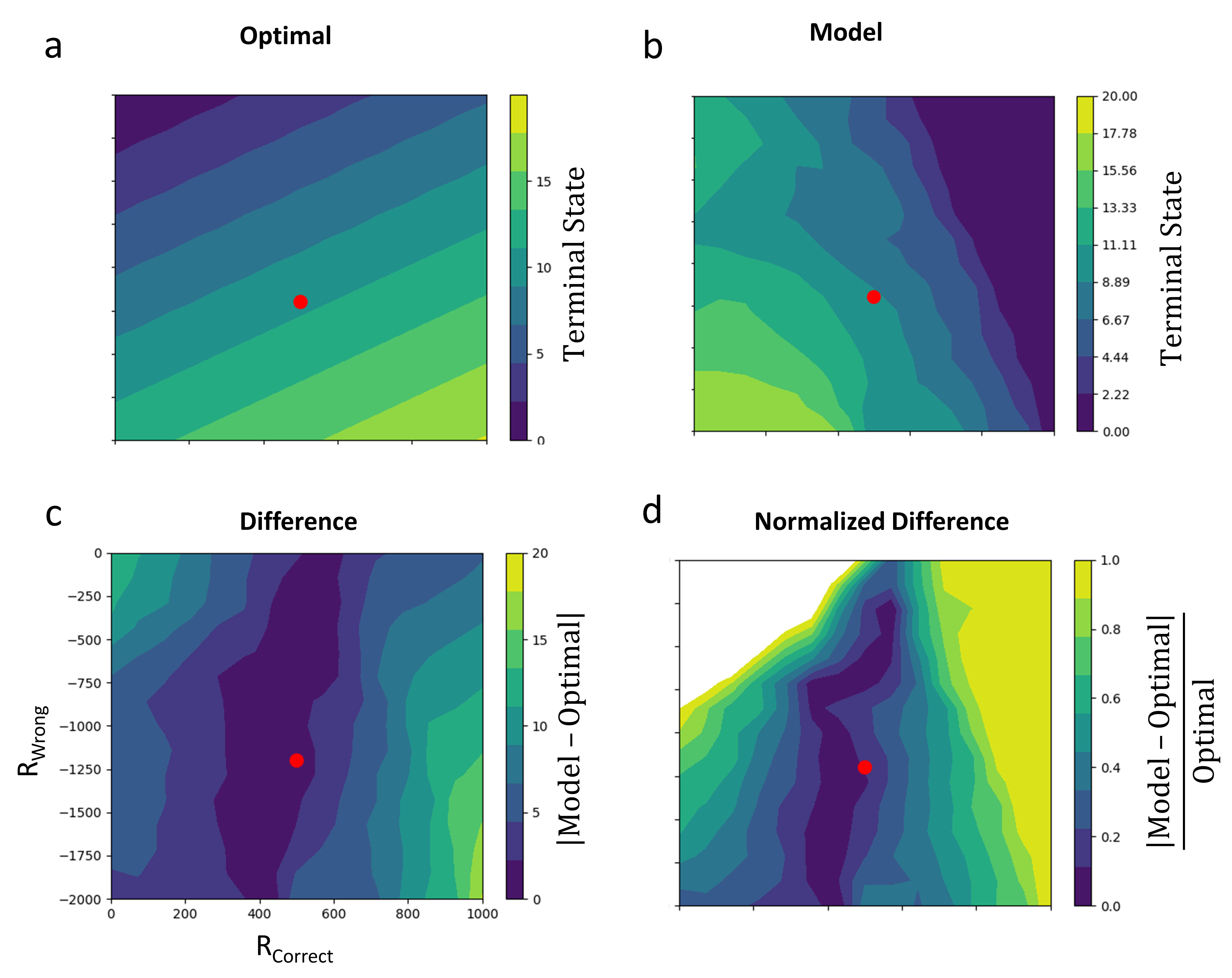}
\caption{\label{fig:oldopt} \textbf{Comparison of our model with the optimal model for$c = 6.4\%$} (a) Optimal terminal state and reward regime for the optimal model Eq.  \eqref{eqn:exp_reward}. We studied the relation of optimal terminal state (colors) to $\Rcorrect$ and $\Rwrong$. The red dot denotes a reward regime that has been used in Fig. \ref{fig:opt}. (b) same as (a) but for models simulations (N=10 iteration, $\epsilon = 0.01, \beta = 50, \Delta = 0.1, \gamma = 1, U = 3000$). (c) Difference between optimal terminal state (a) and model's terminal state (b).  (d) same as (c) but with the normalized difference. The payoff regime that we chose (500, -1200) has a small distance to the optimal model.}
\label{fig:oldopt}
\end{figure}

\section{Comparison with Optimizing Expected Reward}
\label{sec:opt}
Setting a decision threshold on evidence accumulation in perceptual decision making under uncertainty can be thought of as an optimization of a cost function. One reasonable choice for such a function is the expected reward defined as
\begin{equation}  
\label{eqn:exp_reward}
      {\rm ER}(B,K) =  \Rcorrect  {\rm Accuracy}(c|B,K) - \Rwrong  [1-{\rm Accuracy}(c|B,K)] - \Rwait {\rm RT}(c|B,K)
\end{equation}
where ${\rm Accuracy}$, ${\rm RT}$ are the same as those in Eq.\ \eqref{eqn:sat}. 
Since in Eq.\ \eqref{eqn:exp_reward}, $B$ can take continuous values, in the simulations reported in this section, we have also used $\Delta = 0.1$, so as to make the discrimination of the states of the model finer. We consider the case of $\Delta = 1$ which is used in the other results reported so far in \ref{newparams}. We call the value of $B$ for which ${\rm ER}(B,K)$ is maximized as {\it optimal terminal state}. Note that we used the simple grid search over Eq.\ \eqref{eqn:exp_reward} to obtain the optimal $B$.

\begin{figure}[h]
\centering
\includegraphics[width=.95\textwidth]{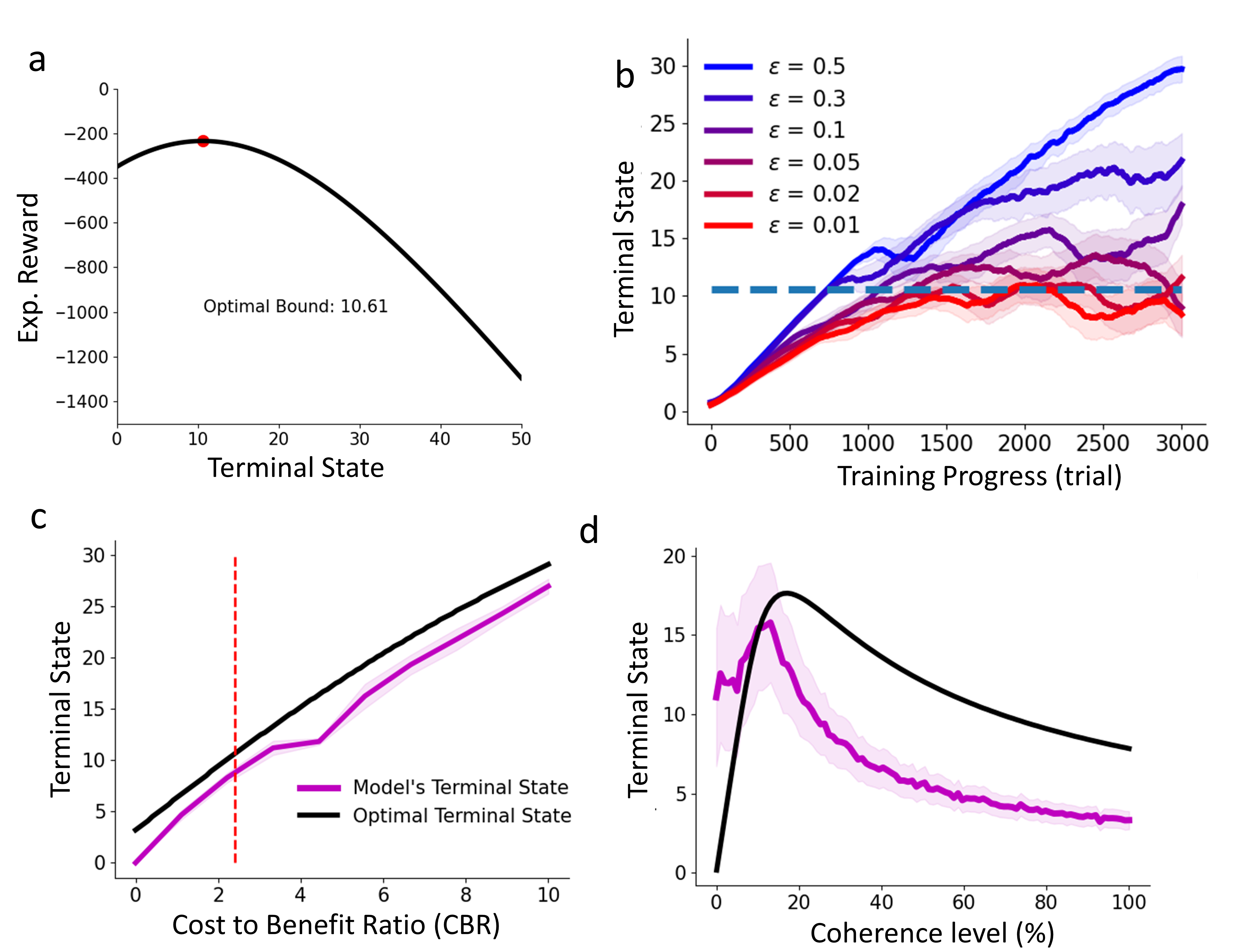}
\caption{{\bf Optimality of the terminal states.} (a) Expected Reward (Eq.
\ \eqref{eqn:exp_reward}) as a function of $B$ for $\bR=[500, -1200, -1], \gamma = 1$ and $c = 0.064$; the maximum is denoted by the red dot. (b) Position of the terminal state reached by the model for different learning rates $\epsilon$, compared to the optimum (dashed line). (c) Optimal terminal states and those reached by the model with  $\epsilon = 0.01$ and after $3000$ trials, plotted versus the cost-benefit ratio (CBR). The model's terminal states in taken as the average of the terminal states over the last 300 trials. CBR corresponding to the parameters in (b) are indicated by a dashed line. (d) Similar to (b) but for different coherence levels; $\epsilon = 0.05$ and $U = 6000$ were used.
In the case of simulations in (b)-(d) the solid curves are averages and the shaded are the STD over 5 simulations. In all simulations here we used $\Delta = 0.1$.}
\label{fig:opt} 
\end{figure}

In Fig. \ref{fig:oldopt}, we show the value of the optimal terminal state (Fig. \ref{fig:oldopt}(a)) and those reached by the model (Fig. \ref{fig:oldopt}(b)), as well as the difference between the two, Fig. \ref{fig:oldopt}(c), for different choices of $\Rcorrect$ and $\Rwrong$ and fixed $\Rwait=-1$. It is clear from these figures, that there is a region in the $(\Rcorrect,\Rwrong)$ plane that the optimal terminal state and that of the model are quiet close to each other (e.g. within $10\%$). Fig. \ref{fig:opt} shows the relationship between the optimal terminal state and that of the model when $\Rcorrect$ and $\Rwrong$ correspond to a point in this region, specifically the point denoted by the red dot in Fig. \ref{fig:oldopt}, with $\bR=\{500,-1200,-1\}$.

In Fig. \ref{fig:opt}(a), we first plot the expected reward as a function of the terminal state $B$; the optimal terminal state is denoted by a red dot. 
Fig. \ref{fig:opt}(b) shows when trained on a fixed coherence $c$, how the terminal state reached by the model compares with the optimal terminal state. For large values of $\epsilon$ (blue curve in Fig. \ref{fig:opt}(b)), the model overshoots the optimal. This means that, across training, the model keeps accumulating more and more evidence, increasing its reaction time, without the accuracy changing significantly: large $\epsilon$ constrains accuracy but not reaction time. The model can, however, arrive at a terminal state close to an optimal one (dashed line) as long as the learning rate $\epsilon$ is adequately small. In sum, the model that is not explicitly designed to optimize the expected reward can indeed be close to optimal in the sense of the cost function in Eq. \eqref{eqn:exp_reward} for some pay-off regimes. 
 
In Fig. \ref{fig:opt}(c) we examine the concordance between the terminal state reached by the model and the optimal one in more detail. Interestingly, when plotted as a function of the cost-benefit ratio, the terminal state found by the the model follows that of the optimal solution closely, with the difference between the two remaining relatively constant as the the cost-benefit ratio changes. Combined with the fact that the actual position of the model and optimal terminal states increase linearly with CBR, for larger CBR, the terminal state of the model can get to only a few percent of the optimal solution; the red dashed line, corresponding to the CBR of the reward values used in Fig. \ref{fig:opt}(b). 

Fig. \ref{fig:opt}(d) shows how changing the coherence level affects the optimal terminal state and the terminal state reached by the model. We can see that although the concordance is not always great, the terminal state reached by the model follows similar trends as the optimal. Firstly, in both cases, the terminal state initially increases with coherence. They both then reach a maximum, before decreasing with $c$. Although the terminal state of the model is consistently smaller than the optimal one for larger $c$ increases, they have comparable slopes of decay with $c$. 

In \ref{newparams} we show that the results described above are also true when we discretize the states of the model with $\Delta = 1$. We discuss the differences between the two choices of $\Delta$ in more detail in that section. We also note that, for the simulations in this section, the coherence level was fixed because the expected reward in Eq.\ \eqref{eqn:exp_reward} was defined for fixed $c$. Extending this definition to a more general case in which $c$ changing during training is possible, but finding the corresponding optimum is a non-trivial task. Yet, the results of Fig. \ref{fig:opt} discussed here indicate that even in such a scenario, the model does not diverge far from the optimal solution. 

\begin{figure}
\centering
\includegraphics[width=0.9\textwidth]{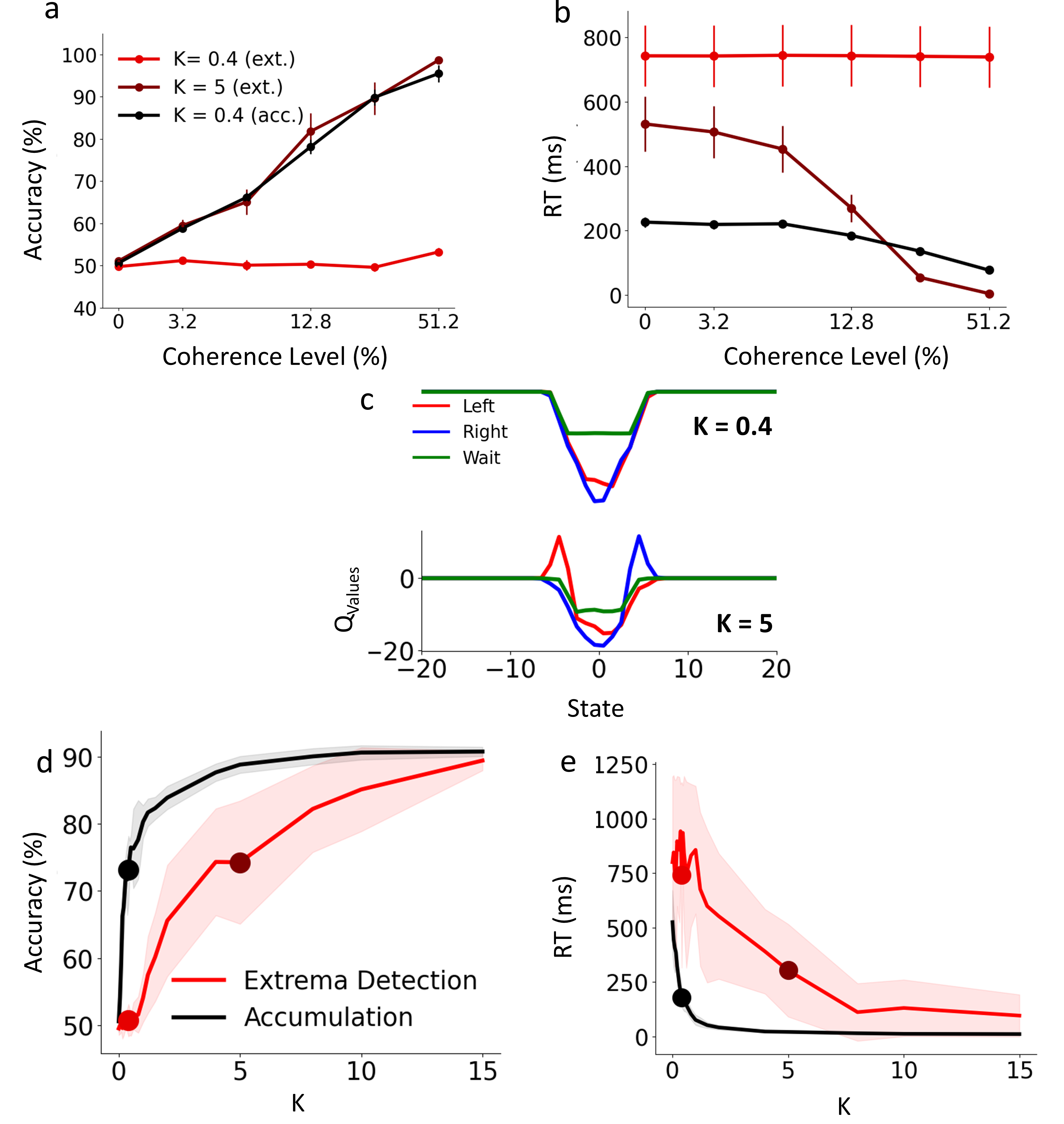}
\caption{{\bf Decision making in the model without accumulating evidence.} (a) The Accuracy of the extrema detection model (Eq.\ \eqref{eqn:state_ext}) with $K=0.4$ (light red) and $K=5$ (dark red) and that of the evidence accumulation model (Eq.\ \eqref{eqn:state}) with $K=0.4$ (black). The extrema detection model with small $K$ (light red) performs near chance level but increasing $K$ leads to accuracy similar, although with more variance, compared to the evidence accumulation model. (b) Similar to (a) but for reaction time. The exterma model with small $K$ stays around the maximum in most simulations (maximum time of stimulus sequence was set to 1000 samples) with few reaching lower values, hence the larger variability. (c) The Q-values after training for the extrema detection model with small $K$ (top) and large $K$ (bottom); compare this to Fig. \ref{fig:trained_model}. (d) Systematic study of the effect of $K$ on the accuracy of extrema detection (red) and evidence accumulation models (black), averaged over all tested coherence levels. The dots correspond to the simulations depicted in (a) using the same color conventions. (e) same as (d) but for RT. Error bars in (a) and (b) and the shaded area in (d) and (e) are SEM across 20 simulations, each including 900 training and test trials.}
\label{fig:ext}
\end{figure}

\section{Deciding without evidence accumulation}
\label{sec:exterma}
Up to this point, our model dispensed with the assumption of a decision boundary. Yet, it still relied on accumulating the moment-to-moment sensory evidence: in Eq.\ \eqref{eqn:state}  $E_t$ is added to $s_t$. Reasonable as it may be, it is important to see whether this operation is a necessary condition for the model to function. In this section, we show that indeed it is not. Even when state dynamics that do not involve accumulation of evidence, action selection and Q-learning, namely, Eq.\ \eqref{eqn:Q-table}) and Eq.\ \eqref{eqn:softmax}, are sufficient for decision making. 

We set the dynamics of the state variable to follow {\it exterma detection} \cite{stine_differentiating_2020, watson_probability_1979}, and we have
\begin{equation}  
\label{eqn:state_ext}
      s^u_t = \nint{E_t}_{\cS}.
\end{equation}

An example of the resulting dynamics is shown in Fig. \ref{fig:ext}, with $\epsilon= 0.1$, reward set $\bR=[20, -50, -1]$ and $M=100$ and the resolution of state space, as before, is set to $\Delta = 1$. Starting with the same parameters with which the evidence accumulation (Eq. \eqref{eqn:state}) model yielded reasonable performance previously, we see (Fig. \ref{fig:ext}(a)-(b)) that, the success of Eq. \eqref{eqn:state_ext} depends critically on the value of sensory gain parameter $K$. When $K$ value is in the range used in the evidence accumulating simulations ($K=0.4$), then $E_t$ in  Eq. \eqref{eqn:state_ext} is rarely sufficiently large to move the model towards higher states. Consequently, for such a small $K$, the system practically gets stuck in a few states around zero. This change, though, when we consider larger $K$ and the model performs satisfactorily. 

Increasing $K$ leads to larger values of $E_t$ for the same coherence level $c^{\mu}$. Even though the model with Eq. \eqref{eqn:state_ext} does not accumulate the momentary evidence it receives, it nonetheless achieves very reasonable performance if we set (for example) $K=5$: the psychometric and chronometric curves shown in  Fig. \ref{fig:ext}(a) and (b), respectively, now reflect the same canonical features that we previously observed in Fig. \ref{fig:trained_model} and in empirical studies. Note that although the accuracy of the extrema detection model matches that of the accumulation model, the reaction time (RT) is consistently longer for extrema detection, especially at lower coherence levels. As can be seen by comparing Fig. \ref{fig:ext}(c) and e.g. Fig. \ref{fig:trained_model}(d), the Q-values associated with the actions in each state after training show the same prominent feature, namely, that terminating states develop on the left and right sides of $0$, corresponding to (correct) Left and Right actions. With Eq. \eqref{eqn:state_ext}, the Q-values of the actions, however, depend more strongly on the states compared to Eq. \eqref{eqn:state}. 

To compare the the state dynamics of the two approaches quantitatively, we kept all the parameters including the trial sequence identical between the two and simulated 20 simulations for each case. Fig. \ref{fig:ext}(d), demonstrates that the accuracy of the accumulation model is higher than that of the extrema detection for all values of $K$. Variability across simulation runs is also larger for extrema detection. For large $K$, however, these differences diminish. Similarly, Fig. \ref{fig:ext}(e) shows that the RT of the accumulation model, for all values of $K$, is consistently shorter than those of the extrema detection although this difference shrinks significantly with higher values of $K$. As a side note for interested readers, increasing the resolution of states by decreasing $\Delta$ can produce similar results. This is not depicted here but made available for examination in the shared code.  

\section{Discussion} 
\label{sec:discussion}
Sequential sampling with integration to bound models has been extensively influential in our understanding of decision making under uncertainty \cite{ratcliff_diffusion_2016, moran_optimal_2015}. These models, however, have not produced a compelling explanation for two crucial issues. First, it is not clear how, when the animals are introduced to a new task, they learn the decision boundary given that they know very little about the experiment and the requirements of the task. Second, the way these models explain how agents adapt to changes in context is not consistent with empirical evidence. Studies of speed-accuracy trade-off in humans and other animals have shown that once the agent has learned the task and performs adequately, they can adapt to the new context. accumulation to bound models explain this by proposing a change in the decision bound, but empirical investigations in primate brains have not found evidence for such a proposal \cite{heitz_neural_2012,kiani_choice_2014,heitz_speed-accuracy_2014}. By introducing a new, simple Q-learning model we addressed both of these issues. 

Two key innovations distinguish our Q-learning algorithm: an action repertoire with an extra, $Wait$ alternative and a state variable that accumulated noisy samples of information from the environment.  The combination of these two innovations allowed the model to learn to decide via sequential sampling without, at any point, comparing the accumulated evidence with a decision boundary. The resulting model provides clear answers to the two outlined above. Regarding the first issue, that learning to decide, our model goes beyond earlier works that used model-based approaches for the determination of boundaries in perceptual decision making under uncertainty \cite{drugowitsch_cost_2012, bogacz_humans_2010}. Departing from those earlier works, our model-free approach does not require detailed explicit knowledge of the underlying statistical structure of the task. The model qualitatively reproduces the canonical hallmarks of chronometric and psychometric behavior observed in perceptual decision making under uncertainty. Earlier applications of model-free RL to perceptual decision making \cite{law_reinforcement_2009, kahnt_perceptual_2011} did not account for reaction times. By naturally accounting for reaction times, our model thus goes one step beyond those earlier models.

Furthermore, our model performed reasonably even \textit{without accumulating} evidence showing that sequential sampling is sufficient, by itself, for perceptual decision making under uncertainty. As long as the internal states are arranged such that larger coherence levels push the model to states further away from zero and do so more rapidly than lower coherence levels, psychophysically plausible decision making needs neither boundary nor accumulation. The key challenge of reinforcement learning here, therefore, is to $exploit$ that correspondence between the external noisy sensory information and internal states.  How that correspondence is implemented in the agent's brain could be a combination of evolutionary, genetic, and developmental processes. This is a fundamental question that is beyond the scope of the work presented here but calls for revisiting the tenets of the commonly held beliefs about the neural mechanism underlying decision making under uncertainty.

Regarding the second issue, that of context, by studying the effect of CBR, we demonstrated that our model could adapt and trade off speed with an accuracy consistent with the requirements of the payoff table: asymmetric payoff tables with a higher reward for correct or higher punishment for incorrect decisions shifted the model's psychometric and chronometric functions consistent with empirical observations \cite{bogacz_humans_2010}. Moreover, when rightwards and leftward movement stimuli were not present with equal probability, the Q-table dynamically changed to adapt to the new distribution producing faster reaction times for the more likely option. Our model thus successfully adapts to changes in both payoff and probability contexts.


A number of previous studies have utilized evidence accumulation models to investigate the dynamics of value-based choice in reinforcement learning \cite{pedersen_drift_2017, frank_fmri_2015, fontanesi_decomposing_2019}. 
It is important to highlight the differences between our work and those previous approaches. In short, whereas they relied on sequential sampling models to characterize the within-trial dynamics of choice in an RL task, we relied on RL to characterize how the sequential sampling process evolves within and across trials.  In those previous works, the RL agent's aim is to find out which choice option (e.g., "bandit") yields the higher reward by iteratively choosing among the options, comparing the outcome to its expectations, and updating its options' values. At any given trial, the choice is made stochastically through a diffusion process (towards a predefined boundary) whose drift rate is a function of the latest update of the option values. In these models, the rate of evidence accumulation is determined by value learning. Here we opted for a much simpler approach by adopting a constant drift rate (c.f. Eq.\ \eqref{eqn:state}) and dispensing altogether with boundary, regulating the evidence accumulation process by model-free RL instead. Integrating the concurrent learning of the drift rate into our model could be a promising avenue for future research. 

A key strength of the sequential sampling framework is that it treats decision making under uncertainty as an optimization problem whose aim is to find the decision boundary that maximizes the reward rate given a combination of signal coherence, payoff regime, average accuracy, and average reaction time. Our simulations revealed that the position of the terminal state in the Q-table in some cases matches that of the optimal solution, and in general the way it changes with the parameters of the task, reflects features similar to those of the optimal solution. Optimizing the expected reward could, of course, be a consequence of Q-learning, but to meet the sufficient conditions for this to be the case, each state-action pair must be tried "many times," and the learning rates must be allowed to decay (see \cite{watkins_q-learning_1992}). However, our simulations do not meet these conditions: the model does not explore all state-action pairs and uses a constant learning rate. Even if these conditions are met, convergence to optimality can take an extremely long time due to the complex nature and large expanse of the state space. This may explain the consistent underestimation of the decision threshold observed in model simulations. 
It is thus encouraging to see that while our model is not geared to any of these requirements it still behaves consistently with the optimal solution. 

A longstanding tradition in the investigation of perceptual decision making is to have subjects (be they human or non-human animals) undergo extensive training to reach a stable, asymptotic level of performance before they participate in the experiment proper. In this tradition, which includes many of the current authors' previous works too, hundreds and thousands of such training trials are discarded because these data (behavioural and/or physiological) are deemed too unstable for interpretation. Another justification for discarding training data was that, up to now, there was no simple and general model for interpreting behavior during learning in perceptual decisions making under uncertainty.  Future studies will be able to use our model to revisit those discarded data and compare them to model predictions.




\printbibliography

\newpage
\section*{Supplementary Material}
\setcounter{subsection}{0}
\renewcommand{\thesection}{S}
\setcounter{figure}{0}
\renewcommand\thefigure{\thesection.\arabic{figure}}

\subsection{A toy example scenario}
\label{sec:toy}
To gain some insight into why the emergence of the terminal states discussed in section \ref{sec:q_during} makes sense, it is imperative to consider the simplest version of the model: when at each trial $u$, the coherence level $c^{u}$ can be either +1 or -1 and $\sigma=0$, $\epsilon = 1$, $\gamma = 0$ and $\beta = \infty$. In this case, the Q-table evolves deterministically and the actions are also taken deterministically, thus making understanding the evolution of the Q-values considerably easier. 
As will be shown below, in this case after a number of trials the Q-table converges to a configuration, whereby in state $0$ the Wait action will have the largest Q-value while in states -1 and 1, Left and Right actions will have the largest Q-values, respectively. Starting from state, 0, the agent will start with Wait then move to state 1 or state -1 depending on the coherence level, and then respectively choose the Right or the Left action. 

\begin{figure}
\centering
\includegraphics[width=0.8\textwidth]{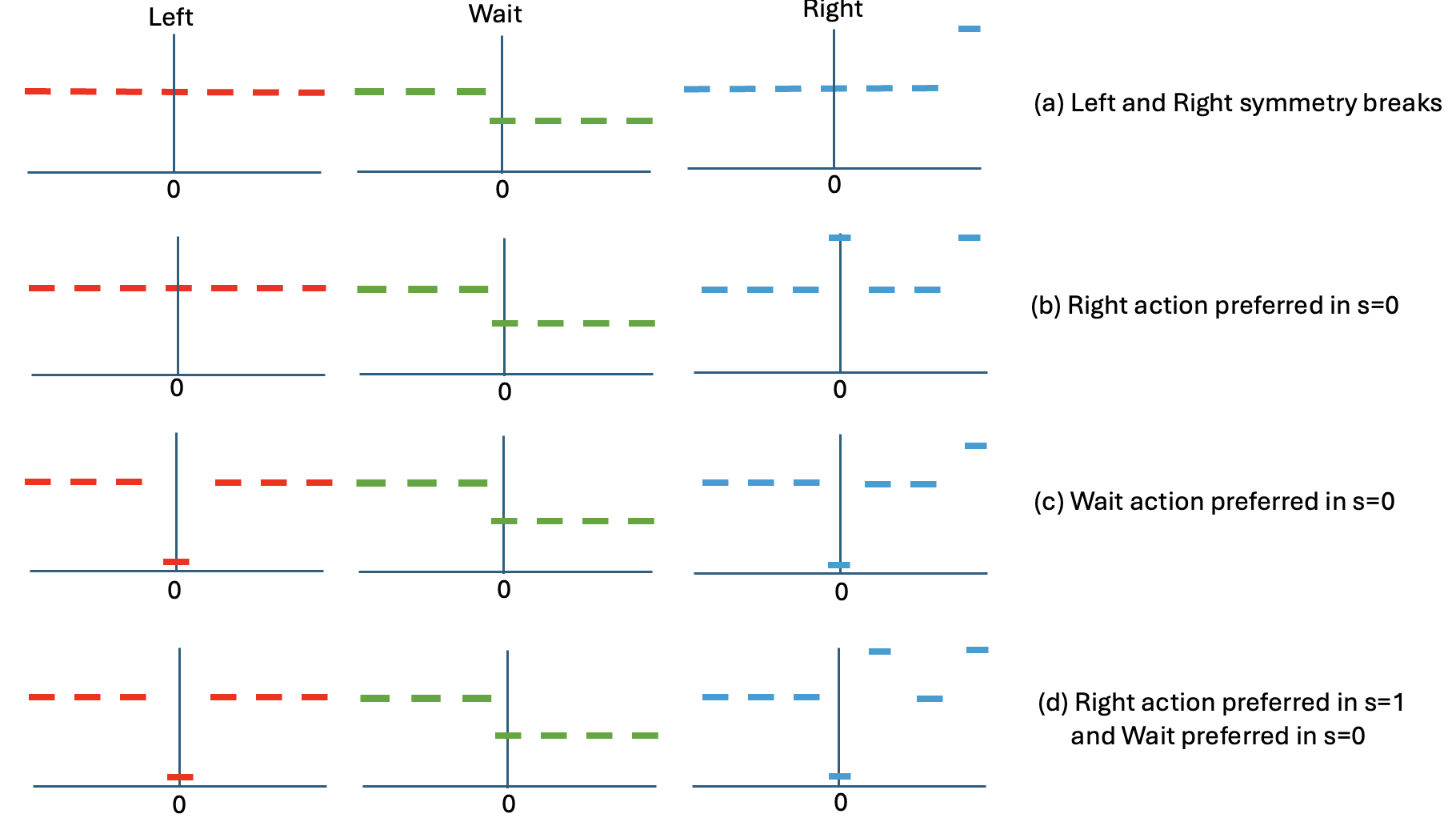}
\caption{\label{fig:toy} \textbf{Cartoon depicting different stages of the Q-table for the toy model.} The Q-values of the different actions are shown for each state for each of the stages of the dynamics discussed in the text.}
\end{figure}

{\bf The Left/Right Symmetry of the Q-table breaks.} Without loss of generality, suppose $c^{1}=1$. The agent makes a Right or Left actions by chance after a number, $\Gamma$, of Wait actions. Since the Q-values are initialized at zero, this happens with probability $(1/3)^{\Gamma+1}$ and the trial ends with the agent in state $s_\Gamma=\Gamma$. Assuming $\Gamma \ge 1$, at the end of the first trial, we have $Q(s,:) =(0,0, \Rwait)$ for $s=0,\cdots, \Gamma-1$. For $s_{\Gamma}=\Gamma$ at which the terminating action is taken, two possibilities exist: $Q(\Gamma, :)=(\Rcorrect,0,0)$ with probability $1/2$ when the final action is correct (Right), or $Q(\Gamma, :)=(0,\Rwrong, 0)$ with probability $1/2$, when it is incorrect (Left). In both cases the $Right$ action ends up being the preferred action in $s=\Gamma$. The important point now is that the symmetry between between $s>0$ and $s<0$ is now broken, as a consequence of performing the first trial. This is shown in Fig. \ref{fig:toy}a. 

{\bf One of Left or Right actions becomes the preferred action at $s=0$.} At the beginning of this trial, $Q(0,{\rm Wait})< Q(0,{\rm Left})=Q(0,{\rm Right})=0$, so Left or Right actions will be chosen with equal chance, and the trial ends. If $c^2=1$, taking the Left action is incorrect and turns $Q(0,{\rm Left})=\Rwrong$, while taking the Right actions is correct, turning $Q(0,{\rm Right})=\Rcorrect$. In both cases, the Q-values of the remaining two actions do not change, and, since $\Rwrong<\Rwait<\Rcorrect$, the Right action will have the largest Q-value at state zero; see Fig. \ref{fig:toy}b. Similarly, if $c^2=-1$, then the Left action will have the largest Q-value at state zero. 

At the beginning of the next trial, then, the ${\rm Right}$ action will always be immediately taken. Note that if we had $\Gamma=0$ in Trial $1$, the agent would end up in this situation at the beginning of Trial $u=2$ and the following would ensue. 

{\bf Wait becomes the preferred action at $s=0$.} Suppose that $c^2=1$.  If $c^3=1$, this will be the correct decision, and $Q(0,{\rm Right})$ will further increase. This continues to be the case in all subsequent trials until a trial with $c=-1$ is encountered. When the first $c=-1$ trial is observed, the Right decision that the agent takes in state zero is incorrect, and thus $Q(0,{\rm Right})$ decreases. After a number of trials, which depends on how many of the preceding trials had $c=1$, as well as how large $|\Rwrong|-|\Rcorrect|$ is, eventually, $Q(0,{\rm Right})<Q(0,{\rm Wait})$. If $c^3=-1$, this happens at the third trial itself. 

In other words, after a sufficient number of trials, not only the symmetry between $s>0$ and $s<0$ is broken ($Q(s,{\rm Wait})<Q(s,{\rm Right})=Q(s,{\rm Left})$ for $s=1,\cdot,\Gamma-1$, and $Q(s,:)=0$ for $s=-\Gamma,\cdots,1$), the Wait action is now the preferred action at $s=0$; see Fig. \ref{fig:toy}c. This is also the case if $c^2=-1$. 
 
{\bf Right action becomes preferred at $s=1$.} In the trial that follows the formation of this structure in the Q-table, the agent first Waits, and if $c=1$, moves to state $+1$. Here it will take Right and Left actions with equal chance leading to $Q(1,{\rm Right}) = \Rcorrect >Q(1,{\rm Left})>Q(1,{\rm Wait})$, or $Q(1,{\rm Right}) =0>Q(1,{\rm Wait})>Q(1,{\rm Left})=\Rwrong$: the Right action will end up having the largest Q-value in state $s=1$; see Fig. \ref{fig:toy}d. If $c=-1$, the agent moves to state $-1$, where all actions have Q-value zero, and then everything will be similar to the case of Trial $u=1$ above but with Right and Left actions reversed. Eventually, the Left action becomes the preferred action in state $-1$. 

Given a sufficient number of trials with right- or left-wards stimuli with similar frequency, converges to a Q-table where the largest Q-values for the -1, 0, and 1 correspond to those of the Left, Wait, and Right actions, respectively. 

\subsection{Effect of the Learning Rate $\epsilon$ and number of states $M$}
\label{sec:stlr}
\begin{figure}
\centering
\includegraphics[width=1\textwidth]{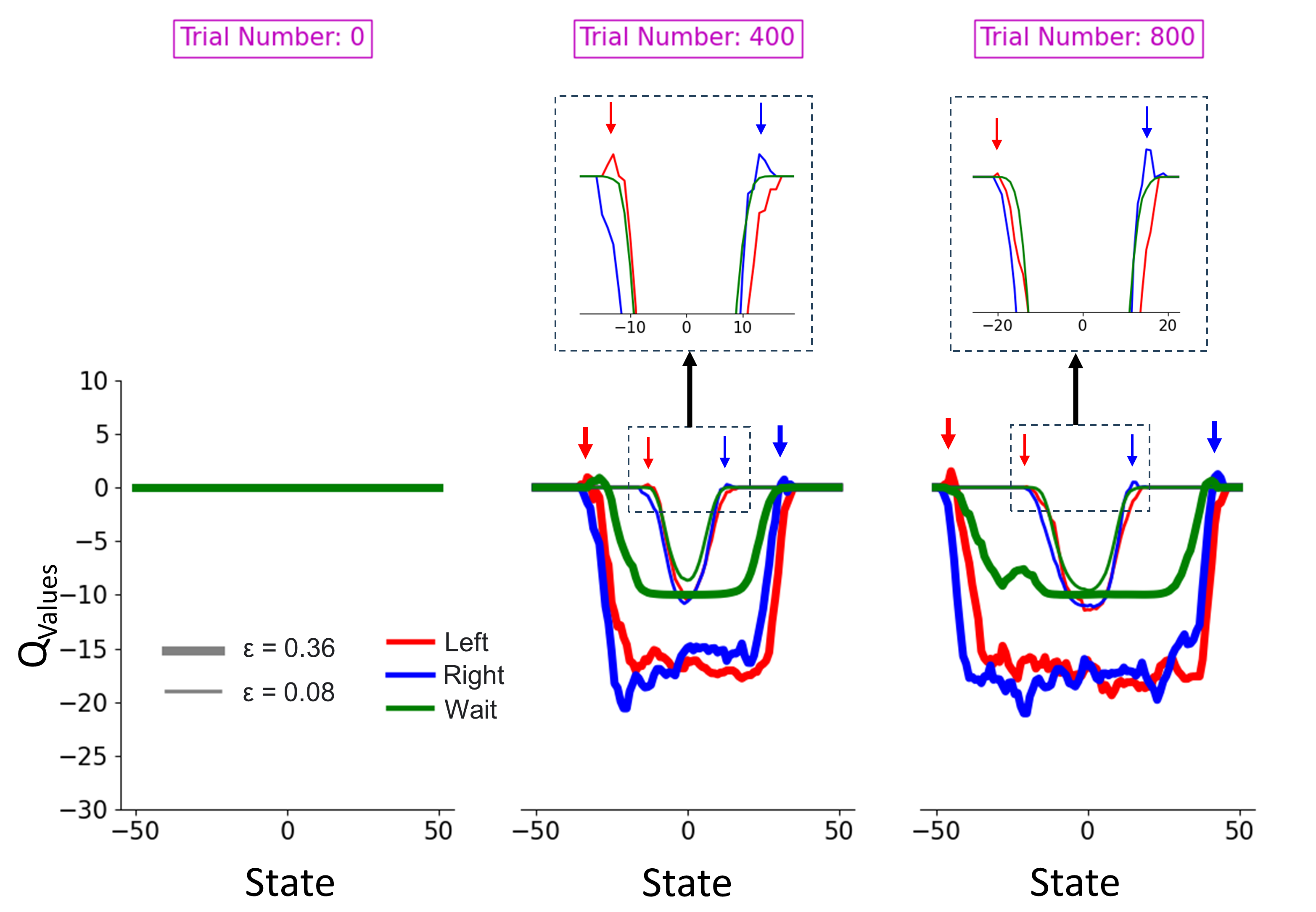}
\caption{\label{fig:lr_sup} \textbf{The effect of learning rate on q-values during learning}. The thick lines are for $\epsilon = 0.36$. The thin lines indicate q-values under $\epsilon = 0.08$ (other parameters are kept the same). For clarity of terminal state positions, the zoomed-in versions are depicted on top of the middle and right panels. For a given number of training trials, larger $\epsilon$ spans wider over state space. Model parameters are $U = 900$.}
\end{figure}

\begin{figure}
\centering
\includegraphics[width=1\textwidth]{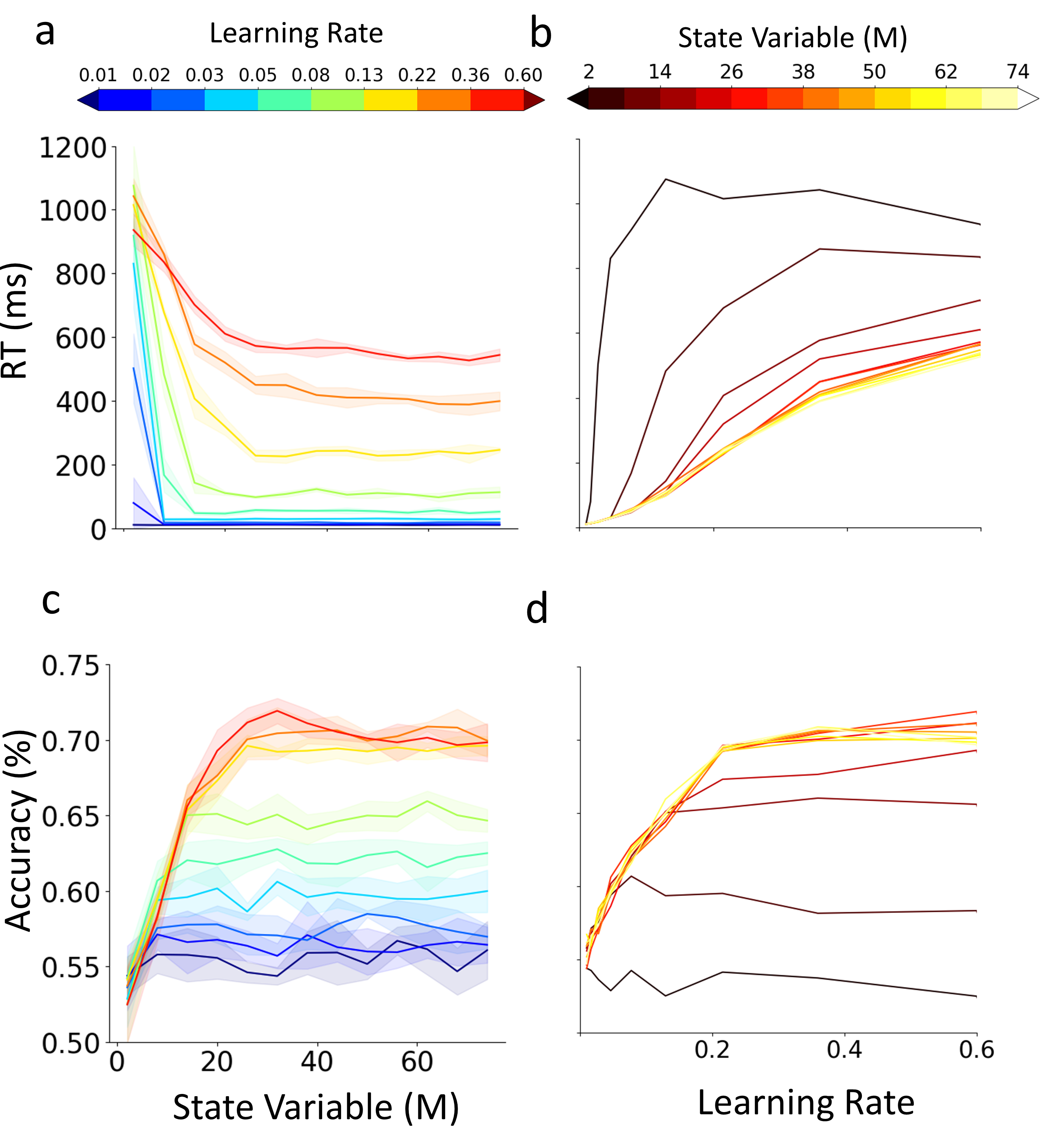}
\caption{\label{fig:lr} \textbf{Effect of learning rate and the available number of states on Model's behaviour (RT and Accuracy) during training }  (a) Average RT (y axis) decreases with increasing the number of available states (x axis). The exact shape of this relationship depends on the learning rate (color bar). (b) Average RT (y axis) does not have a monotonic relationship to learning rates (x-axis). Here, colors represent the number of available states. (c) Same as (a) but for accuracy. (d) Same as (b) but for accuracy. In both (a) and (c), the shaded areas are standard deviations over interactions (n=5). Here, we used a version of the model with no TD and with hard max (namely, $\gamma=0, \beta=\infty$)}
\end{figure}

 The trajectory that Q-values take during training and consequently the development of the terminal states shown in Fig. \ \ref{fig:lr_q} (see section \ref{sec:q_during}) is of course dependent on the learning rate, $\epsilon$. Fig. \ \ref{fig:lr_sup} this dependence, showing that for the model with a higher learning rate, the states that become the terminal states are further to the right or left of those with smaller learning rates, of course, is such states exist. Naturally, then, the accuracy and reaction time achieved by the model after a fixed number of trials depends on both $M$ and $\epsilon$ as shown in Fig. \ref{fig:lr} for $U=900$. For any given learning rate, $\epsilon$, average reaction time (Fig. \ref{fig:lr}(a)) and average accuracy (Fig. \ref{fig:lr}(c)) quickly asymptotic as $M$ is increased. Fig. \ref{fig:lr}(b)-(d) show the same data re-plotted with the learning rate, $\epsilon$, on the x-axis and the number of states $M$ determining the color code. For very small values of $M$, average reaction times show a nonlinear relationship to the learning rate and accuracy remains very low. As $M$ is increased to $\gtrsim 40$, however, average reaction time shows progressively more linear relationship to the learning rate.

\subsection{SAT During learning: Observation Learning and Time Dependent Waiting}
As discussed in section \ref{sec:training}, when the Q-values are all initialized 
at zero, the reaction times at the beginning of training are small while they increase as training progresses. This pattern is not consistent with several previous empirical studies and in this section, we propose two alternative ways to resolve this inconsistency.

\label{sec:stlr}
\subsubsection{Observational learning}
To explain this inconsistency between empirical observations and theoretical observations, we reiterate that our RL model starts with a blank slate, i.e., without any experience of what the stimulus consists of and how it relates to action and outcome. This is not the case in empirical studies of decision making neither in human nor in animal experiments. Human participants are often given a set of “instructions” explaining to them, often verbally, what they are about to experience and what they are expected to do. Moreover, in nearly all psychophysical experiments, before starting the experiment, participants get a few “warm up” trials in which the experimenter demonstrates the stimuli, the correct choices, the actions, and the corresponding rewards. In animal studies too, experiments are preceded by extensive and elaborate acclimatization and habituation procedures to familiarize the animal with the environment and its affordances such as the stimuli, and the available actions. These procedures are necessary to reduce the animal’s anxiety in the novel environment and to draw its attention to the relevant components of the experimental setup. These procedures are often specific to the lab, animal species, experimental question, and the experimenter’s personal style and experience. The impact of these informal procedures, both in humans as well as other animals, on behavior is unknown and often ignored under the convenient assumption that the scientific question of the study must be more interesting and important than issues as trivial as the impact of instructions. Indeed, our own formulation of the RL model described so far did not take this issue into account. To close this gap, we simulate the instructions that human participants receive in psychophysical experiments on decision making by giving the model a short “warm-up” with observational learning before embarking on its instrumental learning task. 

\begin{figure}
\centering
\includegraphics[width=1\textwidth]{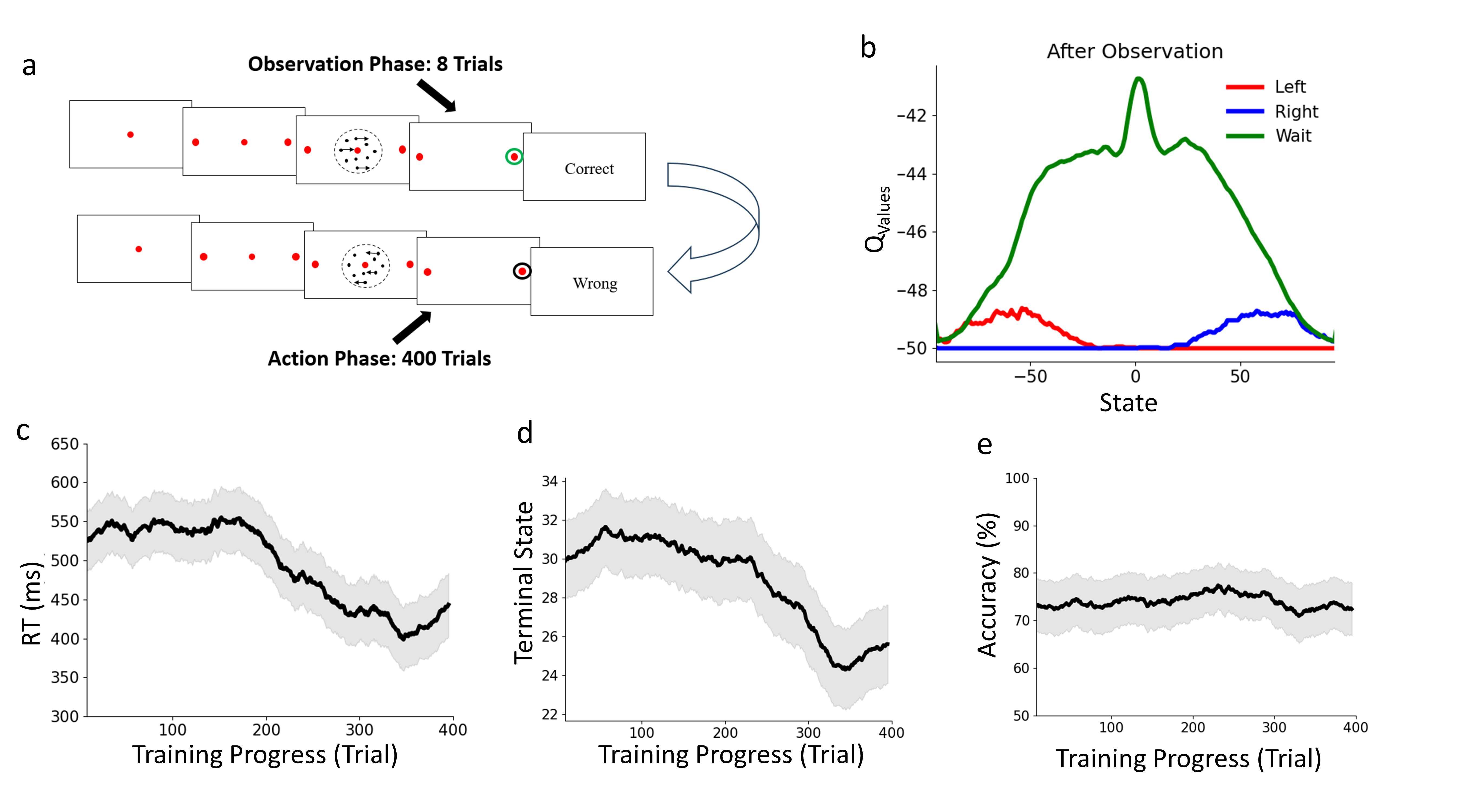}
\caption{{\bf Model behaviour following a period of observational learning.} (a) Observational Learning paradigm. Following \cite{burke_neural_2010} the model {\em observes} the stimulus and the behaviour of a demonstrator in 8 trials. In each trial, a highly coherent stimulus (51.2\%) is presented for a fixed duration of 300 time steps followed by the execution of the correct action and reception of the corresponding reward. The model updates its state and Q-table accordingly. (b) Q-value at the end of observational learning, averaged over 100 simulations.  The Q-values at the end of the observational learning is used as the initial condition for the model going through 400 trials of learning similar to those the main text (Eqs.\ \eqref{eqn:state}, \eqref{eqn:softmax}, \eqref{eqn:Q-table}). (c) The reaction times, (d) terminal states and (e) choice Accuracy as learning progresses. In (c)-(e) Solid curves depicts averages over 50 simulations, each smoothed as in Fig. \ref{fig:BehavebyTime}, and grey areas indicate SEM. The rewards set used for this analysis is $\bR=\{100,-50,-1\}$.}
\label{fig:obs_sup}
\end{figure}

In addition to learning by doing, one can acquire new behaviors, skills, or information by watching the actions and outcomes of others. In the RL framework, the observational learner can update its value function (for example, the Q-table in our formulation here) by observing a demonstrator’s chosen actions and their outcomes \cite{burke_neural_2010, najar_actions_2020, charpentier_neuro-computational_2020}. We had the model with the blank slate initial Q-table observe a demonstrator going through several demonstration trials of the random dot motion task Fig. \ref{fig:obs_sup}(a). In each trial, the stimulus had high coherence (\%51.2). The demonstrator chose the Wait action at consecutive time steps of the trial while the observing learner updated its state variable and Q-table. When the stimulus duration came to its end, the demonstrator chose the correct terminal action and received the reward $\Rcorrect$, again with the observing learner updating its Q-table. Fig. \ref{fig:obs_sup}(b) shows the observer’s Q-table after 8 demonstration trials, each of them comprised of 400-time steps. This \textit{instructed} model was then tasked with learning to make its own decisions. Here, the model started with long reaction times and became progressively faster without losing accuracy (Fig. \ref{fig:obs_sup}(e)), via lowering the terminal state (Fig. \ref{fig:obs_sup}(d)).

\subsubsection{Time Dependant Waiting}
Another plausible solution for the discrepancy between the results of Fig. \ref{fig:BehavebyTime} is making the cost of the Wait action, $\Rwait$ time-dependent, e.g. by using the following
\begin{equation} 
\label{eqn:twait}
  \Rwait(u) = \frac{\Rwait^{\infty}}{1 + e^{\lambda (\tau - u)}}
\end{equation}
where $\Rwait^{\infty}$ defines the asymptotic value of $\Rwait(u)$ as time approaches infinity, $\lambda$ determines how quickly $\Rwait(u)$ approaches this value and $\tau$ determines the point at which $\Rwait(u)$ is halfway to $\Rwait^{\infty}$. 

\begin{figure}
\centering
\includegraphics[width=1\textwidth]{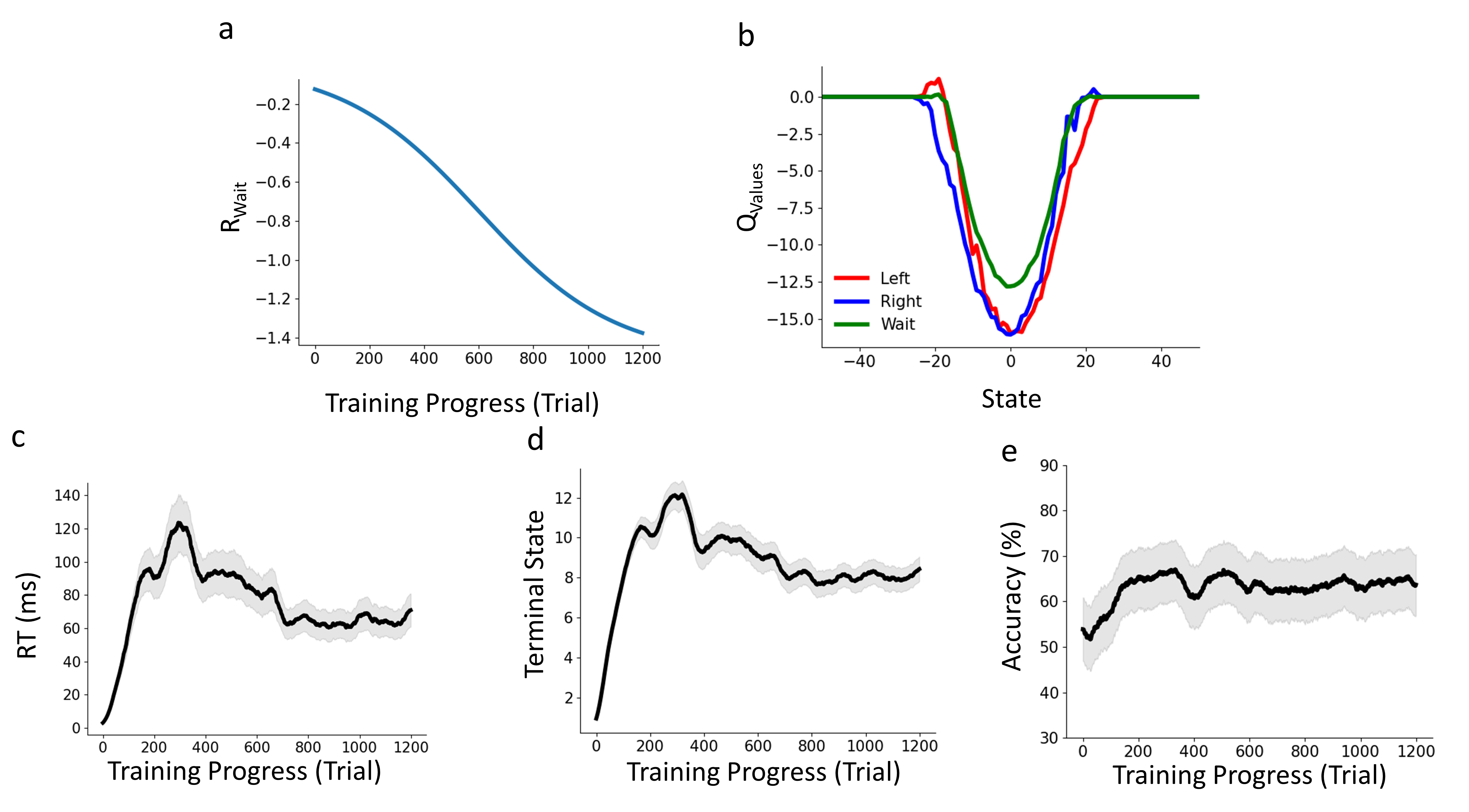}
\caption{ {\bf Effect of time-dependant cost of waiting}. (a) Cost of waiting increases (that is $\Rwait<0$ decreases) as trials progress (Eq.\ \eqref{eqn:twait}). (b)-(e) Same as Fig. \ref{fig:obs_sup}(b)-(e) but for the the case time-dependent $\Rwait$ and and 50 simulations.}
\label{fig:wait_sup}
\end{figure}

In the simulations in this section, we used $\Rwait^{\infty}=-1.5$, $\lambda = 0.004$ and $\tau=600$, resulting in $\Rwait(u)$ as plotted in Fig. \ref{fig:wait_sup}(a) with the cost of waiting at the start of the experiment to be $\Rwait(0) = -0.12$.

In the beginning, since the cost of waiting is not significant, the model tends to move toward higher states as lower states are perceived as less rewarding. This results in an increase in both response time (RT) and accuracy. This behavior is also typically observed in the base model (see Fig. \ref{fig:BehavebyTime}). However, as the cost of waiting progressively increases (Fig. \ref{fig:wait_sup}(a)), the model ceases to advance to larger states. Instead, it is enforced to make decisions more quickly by reducing the terminal state. Consequently, after an initial rise, the model begins to decide faster (see Fig. \ref{fig:wait_sup}(c)). This reduction in the terminal state, however, is not substantial enough to affect accuracy. Therefore, after an initial rise, accuracy remains unchanged (see Fig. \ref{fig:wait_sup}(e)).

\subsection{Comparison with optimal terminal state for $\Delta = 1$}
\label{newparams}
In section \ref{sec:opt} of the main text, we defined optimal terminal state as the value of $B$ that maximizes $ER(B,K)$ in Eq.\ \eqref{eqn:exp_reward} with other parameters fixed. Since Eqs.\ \eqref{eqn:sat}, and Eq.\ \eqref{eqn:exp_reward}, are often defined over a continuous range for $B$, in that section we considered the case of of finer discretization of the state space by assuming $\Delta = 0.1$ in Eq. \eqref{eqn:state}. 

\begin{figure}[h]
\centering
\includegraphics[width=.95\textwidth]{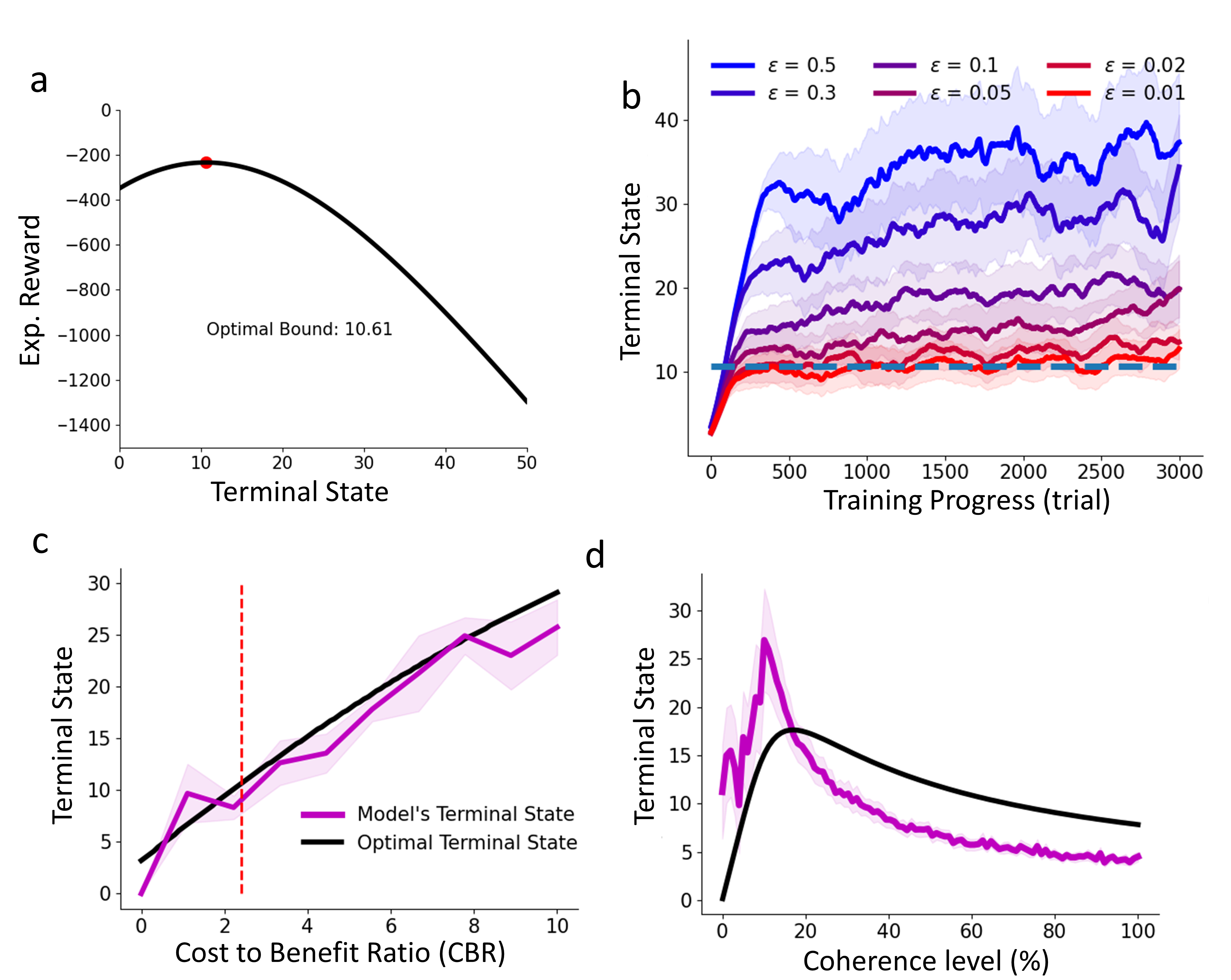}
\caption{{\bf Optimality of the terminal states for $\Delta = 1$.} Everything is the same as Fig. \ref{fig:opt}, except that we now have $\Delta = 1$.}
\label{fig:opt_delta} 
\end{figure}

\begin{figure}
\centering
\includegraphics[width=.9\textwidth]{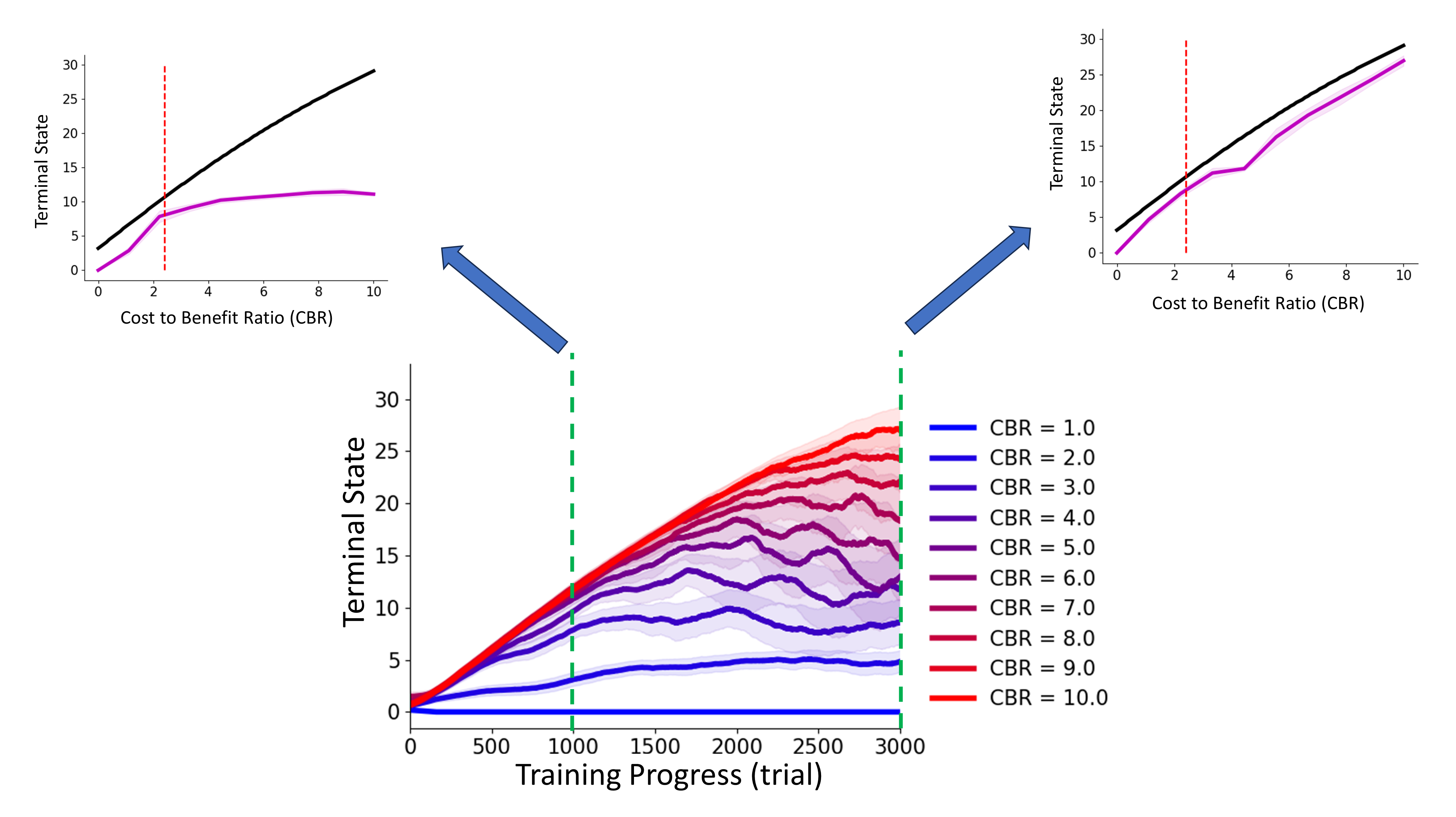}
\caption{\label{fig:opt_trial} \textbf{Sensitivity of the model's optimal terminal state to the number of trials.} If we simulate the model for $U = 1000$ the models lose its sensitivity to high values of CBR (inset, left). Yet, if $U = 3000$ the model will be more sensitive to CBR and show a behavior similar to the optimal model.}

\end{figure}

Here in Fig. \ref{fig:opt_delta}, we show what happens if the discretization is not as fine: we choose $\Delta = 1$, a value that we have used in our other simulations. Everything here is  the same as Fig. \ref{fig:opt}, except for $\Delta$. One can see that, overall, the results do not qualitatively change. 

Quantitatively, however, there are some small differences. Firstly, as expected, the terminal state of the model reaches the optimal after fewer learning trials (Fig. \ref{fig:opt_delta}(b) vs Fig. \ref{fig:opt}(b)). Perhaps more noteworthy is the fact that when plotted versus CBR, the variability in the terminal state reached by different simulations of the model is larger for $\Delta =1$ than $\Delta = 0.1$ (shaded magenta Fig. \ref{fig:opt_delta}(c) vs Fig. \ref{fig:opt}(c) but that the average is still close but mostly smaller than the optimal terminal state. Since, similar to the corresponding figure in the main text (Fig. \ref{fig:opt}(b)-(c)), these results are for $c=0.064$, in Fig. \ref{fig:opt_delta}(d) we also compare the terminal state reached by the model with the optimal terminal state for different coherence levels. Although the overall pattern is the same as Fig. \ref{fig:opt}(d), with $\Delta = 0.1$, the model terminal state typically is at higher values and in fact overshoots that of the optimal states for a range of $c$.


\subsection{The effect of various parameters on the terminal state}
\label{paramstudy}
\begin{figure}
\centering
\includegraphics[width=.99\textwidth]{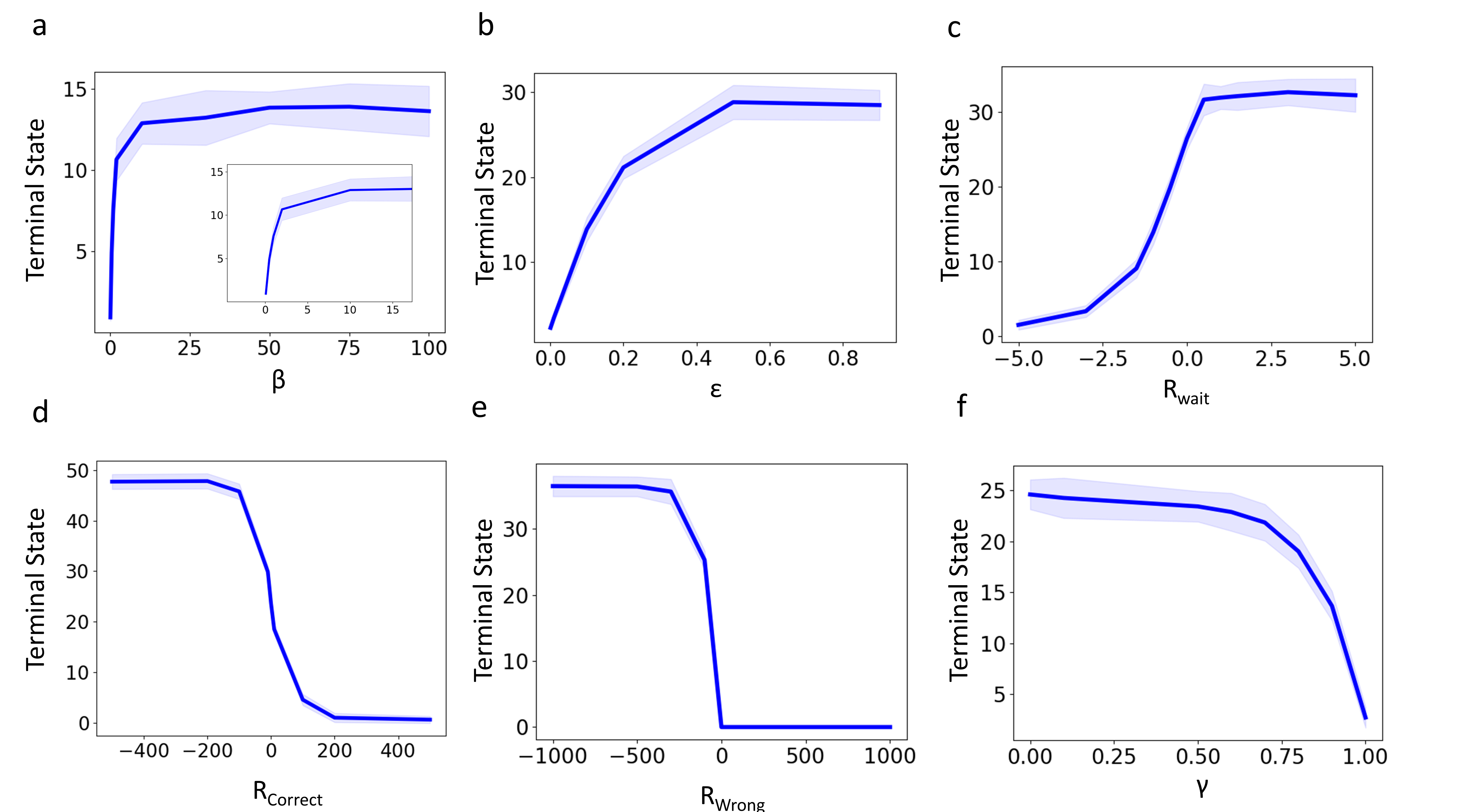}
\caption{\label{fig:paramstudy} \textbf{Systematic study of parameters respect to the terminal state.} The shared parameters that have been used in this study are as follows {$\bR=\{20,--50,-1\}$, $\gamma$ = 0.9, $\beta$ = 50, $\epsilon$ = 0.1, the maximum number of samples per trial = 1000}. In each panel, the parameter of the interest (x-axis) was systematically altered while the other parameters were fixed as above. (a) study of $\beta$ (Eq.  \eqref{eqn:softmax}. The shaded area is STD over simulations (N=30). Inset is the zoomed-in version of (a) for smaller values of $\beta$. (b) Same as (a) but for $\epsilon$ in Eq. \eqref{eqn:Q-table}. (c) Same as (a) but for reward of waiting. (d) Same as (a) but for the reward of being correct. (e) Same as (a) but for reward of being wrong. (f) Same as (a) but for $\gamma$ in Eq.  \eqref{eqn:Q-table}.}

\end{figure}

For the majority of the analyses shown in this work, we used parameter values reported in section \ref{sec:setup}. In Fig.  \ref{fig:paramstudy}, we report a systematic study of the impact of each parameter on the terminal state reached by the model. We used a fixed set of parameters for each analysis while systemically changing the parameter of interest. The number of training trials was fixed at $U = 900$ and the terminal states at the last 100 of these were averaged.


\subsection{Replication of previous empirical findings}
\label{sec:replic}
Similar to any other cognitive computational model, our model could serve as a virtual participant in numerous empirical studies found in the literature of perceptual and value-based decision making and test whether one could "replicate" those findings. Importantly, our model could be tested at the level of its behavior (i.e., reaction time and accuracy) as well as at its "mechanistic" level, for example via examining the model's Q-table structure and the dynamics of its state transitions.

\begin{figure}
\centering
\includegraphics[width=.99\textwidth]{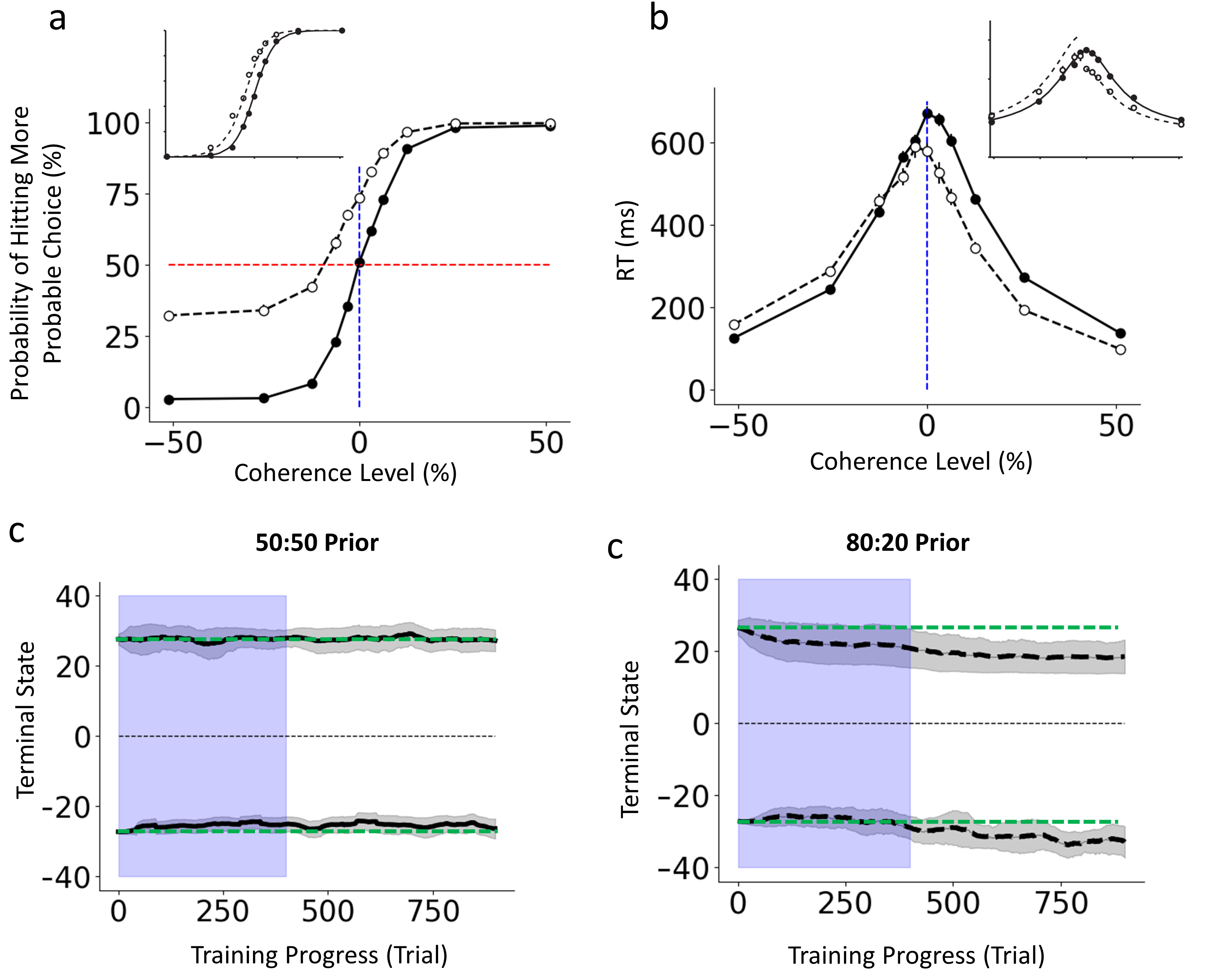}
\caption{\label{fig:dispor} \textbf{The impact of disproportionate training schedule on choice bias and reaction time.} Choice (a) and RT (b) in different training regimes (Solid line: 50-50 prior, dashed line: 80:20 prior, each condition is simulated with 900 trials). Model behaviour becomes more biased (in both decision and RT) toward more frequent (here: rightward) decision under 80:20 regime. Insets are previous findings from by \cite{hanks_elapsed_2011}. Error bars in (a) and (b) are SEM over simulations (N = 30). (c) Evolution of terminal states during 50:50 prior training schedule. The highlighted area is where the data are deemed to be unstable in ref \cite{hanks_elapsed_2011}. Grey-shaded areas are SD over simulations (N = 30). the green dashed line shows the terminal state of the model \textit{before} entering balanced or unbalanced conditions.  (d) same as (c) but for 80:20. Here, the terminal state is significantly lower than the initial terminal state (green dashed line)}.

\end{figure}

Building up from our basic model (Eq. \eqref{eqn:state}-\eqref{eqn:Q-table} and section \ref{sec:trained_model}), we present three successful replications and one important failed but instructive replication attempt. Brefily, we replicated the effects of disproportionate training sets (\cite{hanks_elapsed_2011}, Fig. \ref{fig:dispor}), reverse pulse  (\cite{kiani_choice_2014}, Fig. \ref{fig:pulse}) and volatility ( \cite{zariwala_limits_2013}, Fig. \ref{fig:vol_replicate}). Our basic model could not replicate the effect of differences in RTs for correct vs error trials (Fig. \ref{fig:err_corr}(a)). Motivated by this failure, we introduced a more sophisticated version of the model with an urgency component. This new model replicated the error vs correct RT effects (Fig. \ref{fig:err_corr}(b)) as well as the post-error slowing effect (Fig \ref{fig:PES}) as reported in  \cite{purcell_neural_2016}.
\subsubsection{The impact of disproportionate training set on choice bias and reaction time}
To study the role of prior expectations on perceptual decision-making, Hanks \textit{et al} \cite{hanks_elapsed_2011} examined human and macaque monkey motion discrimination behaviour under two different schedules differentiated by the frequency of stimulus categories. In \textit{balanced} blocks, the two stimulus categories (e.g., leftward and rightward) were equally (50:50) likely. In the \textit{disproportionate} blocks, one motion direction was 4 times more likely than the other (i.e., 80:20). The results showed that in these latter blocks, after 300-400 trials, choices favored the more frequent category (Fig. \ \ref{fig:dispor}(a)-(b) inset) for which RT was also faster. 
\newline
We compared our basic model's behaviour (Eq.  \eqref{eqn:state}, \eqref{eqn:softmax} and \eqref{eqn:Q-table}) under the two schedules (i.e., 50:50 and 80:20). We began by training two instances of the model in the same balanced schedule (3600 trials). Then, one instance went through another 900 trial of learning in the same balanced schedule. The other instance, however, received 900 trials of the disproportionate schedule. 
The results did produce the choice (Fig. \ref{fig:dispor}(a)) and RT bias (Fig. \ref{fig:dispor}(b)) favoring the more frequent category, replicating the empirically reported findings (Fig. \ref{fig:dispor}(a)-(b) inset). Examining the evolution of the terminal states in the two instances (50:50 in Fig. \ref{fig:dispor}.c and 80:20 in Fig. \ \ref{fig:dispor}(d)) showed that after the initial balanced schedule, the terminal states of the two instances were positioned in identical positions and as expected, did not change in the subsequent 900 test trials. In the 80:20 schedule, (Fig. \ \ref{fig:dispor}(d)) however, the terminal state corresponding to the more probable category drifted slowly closer and closer towards zero. The opposite observation was made for the terminal state corresponding to the less probable category. 
\newline
This replication is also important for another related reason. Following the standard practice empirical studies of decision making (see Introduction), Hanks and colleagues discarded the first 300-400 trials of the disproportionate blocks in their experimental data. In these initial trials, the behavior was deemed too unstable for the DDM models to cope with. Here, the results in panels a-b of Fig. \ref{fig:dispor} by follow \cite{hanks_elapsed_2011} and do not include the initial 300 trials of the test phase. But, and here is the important point, the model can also be interrogated during that initial window (the blue areas in Fig. \ref{fig:dispor}(c)-(d)), thus drawing very specific predictions for the very same discarded period. As such, our model gives unprecedented insights into processes that decision neuroscience, especially in animal studies, has so far discarded as uninterpretable and therefore uninteresting. 

\subsubsection{Reverse pulse effect}
\begin{figure}
\centering
\includegraphics[width=.99\textwidth]{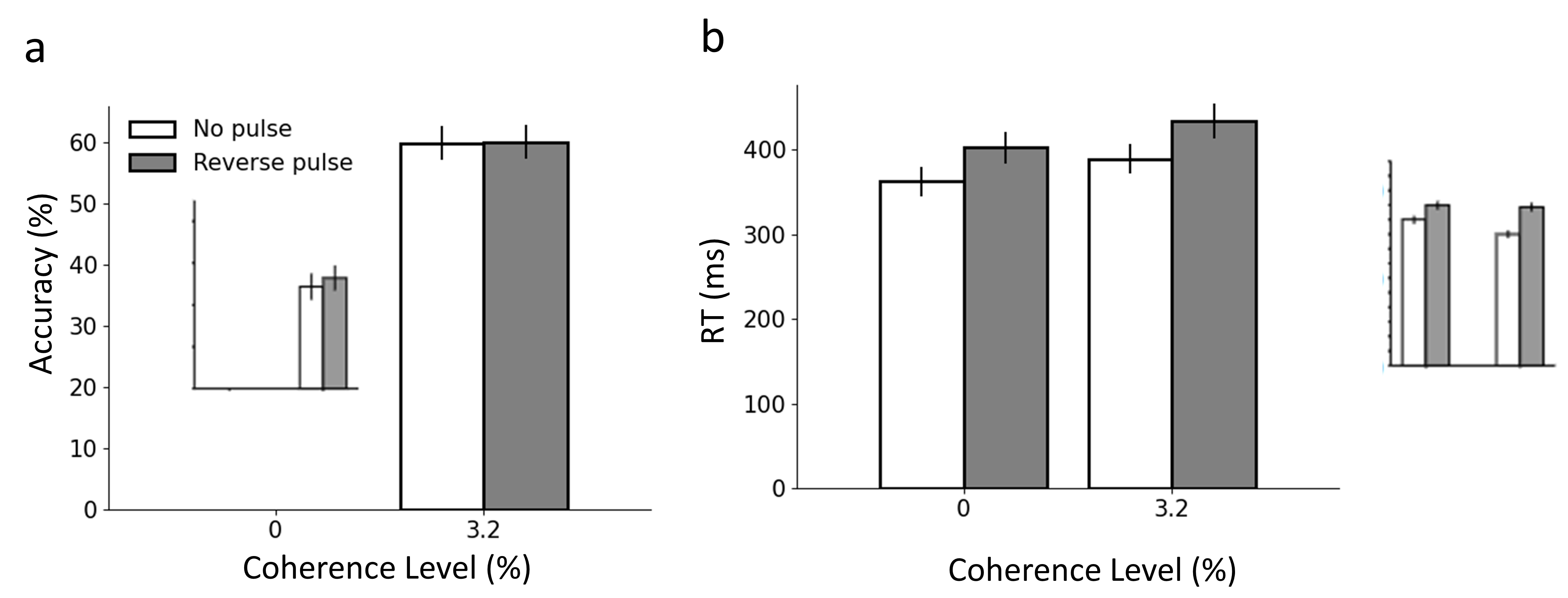}
\caption{\label{fig:pulse} \textbf{Replication of the reverse pulse effect.} (a) Accuracy and (b) reaction times draw from simulation that compared the basic model performance with (grey) and without (white) reverse pulse. Inset show the previous finding from ref. \cite{kiani_choice_2014}. Error bars are 95\% Confidence Interval across trails (U = 2400).}

\end{figure}

The standard drift diffusion model predicts that if a perceptual stimulus is arranged such that statistically \textit{zero evidence} is presented to the observer, the process of evidence accumulation should prolong and decision be postponed such that reaction times increase without any measurable effect on accuracy. This prediction was tested empirically by the so-called \textit{reverse pulse} paradigm \cite{kiani_choice_2014}. Specifically, in this paradigm, at some point during the presentation of the random dot motion stimulus, a brief (200 ms) period  of \textit{zero evidence} is surreptitiously inserted into the motion stimulus. In the first 100ms of this period, the random motion stimulus follows the statistics dictated by the coherence level of the trial. In the second 100ms of this period, motion stimulus was constructed by \textit{reversing} the first half. In this way the motion evidence in the first and second half should cancel the accumulated evidence from one another. Importantly, this trick was only feasible to test in the low coherence trials otherwise the subjects could clearly notice the motion reversal in the middle of the trial. Behavioral results (Fig. \ref{fig:pulse}(b)-(c) insets) confirmed the predictions: insertion of the reverse pulse increased the reaction times but did not affect the accuracy. 

To replicate this finding, after training two identical instances of our basic model (learning rate=0.1; reward set = 20, -50, -1), we tested one instance with and the other without reverse pulse inserted into their stimulus sequence (Fig. \ref{fig:pulse}). Except for the reverse pulse window, the sequences of motion stimuli in the two conditions were carefully matched and coherence was restricted to very low values (0\% and 3.2\%) trials. The trained model (Fig. \ref{fig:pulse}) replicated the empirically observed behavior (Fig. \ref{fig:pulse}(b)-(c)) of longer RTs and no change in accuracy.

\subsubsection{Reaction times in correct and error trials: the role of urgency signal}
\label{sec:urgency}
\begin{figure}
\centering
\includegraphics[width=.99\textwidth]{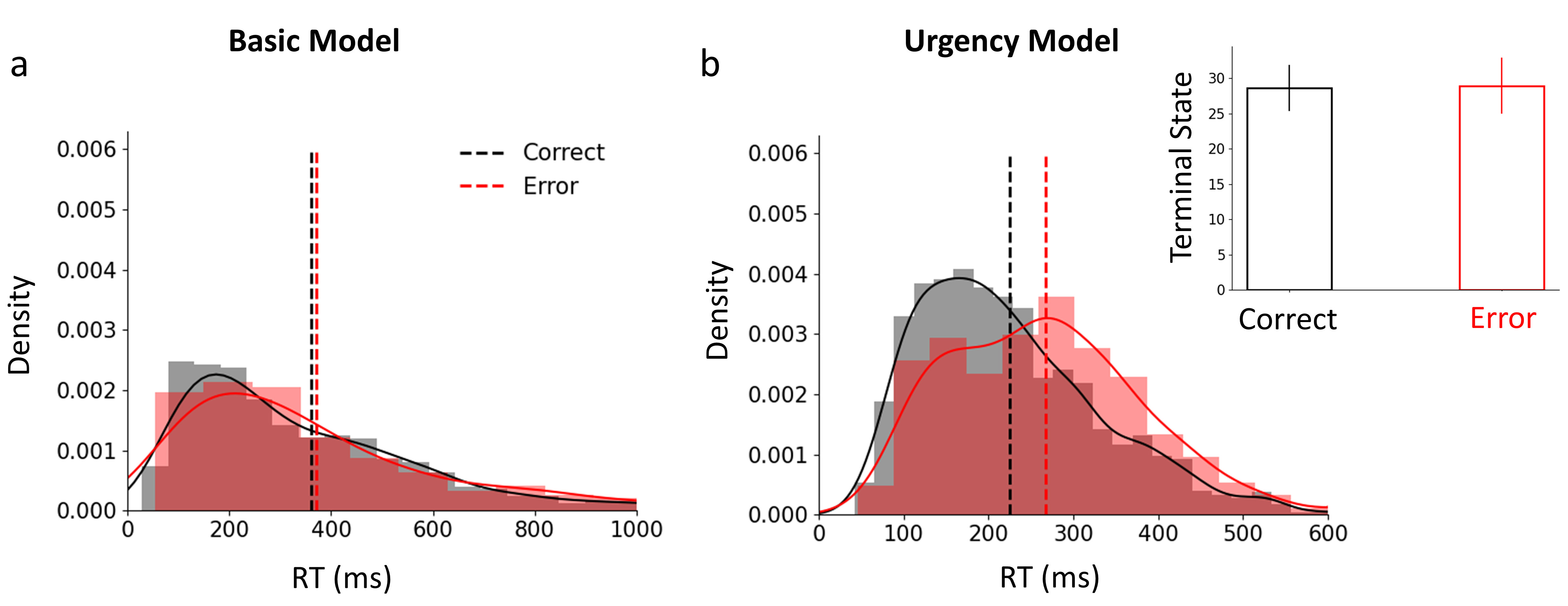}
\caption{\label{fig:err_corr} \textbf{The urgency effect} (a) In the basic model, average error and correct RT are identical. (b) In the urgency model, however, error RTs are longer. The level of coherence is fixed at 12.8\% for this simulation. (inset): Average and STD (error bar) of terminal state for correct (black) and error (red) trials. The difference is not significant (t(2399)=1.5, p=0.13).}
\end{figure}

A common observation in perceptual decision making is that, controlling for task difficulty, error reaction times are longer than those of correct decisions \cite{shadlen_decision_2013}. This intuitive empirical fact has proven to be a critical challenge for many models of decision dynamics. The canonical form of the drift-diffusion model, for example, cannot account for the longer error reaction times \cite{shadlen_decision_2013, ratcliff_theory_1978, cisek_decisions_2009}. Our basic model (Eq.  \eqref{eqn:state}-\eqref{eqn:Q-table}) too is unable to reproduce (Fig. \ref{fig:err_corr}(a)) this widely reported empirical observation. 
\newline
To address this challenge, previous works have proposed a number of different solutions including collapsing boundaries, urgency signal, and drift rate variability. One idea is inspired by the intuition that in the absence of convincing evidence, we may lower our bar for what counts as acceptable \cite{okazawa_neural_2023}. Under high uncertainty, this idea suggests, the boundary come closer to the starting point as time elapses. The longer the evidence accumulation continues, the lower the decision threshold and thus, the higher the probability of erroneous decisions. In simulations, collapsing boundaries could reproduce the difference between correct vs. error RT \cite{okazawa_neural_2023}. Physiological evidence, however, has not supported this proposition. For example, \cite{purcell_neural_2016, roitman_response_2002} showed that the maximum firing rate of decision-boundary neurons in macaque lateral intraparietal (LIP) area do not differ between error and correct trials. An alternative that shares some mathematical similarity to collapsing boundaries is the urgency signal. The intuition behind this idea is that the agent prefers to make faster decisions and when the trial takes longer, some sort of internal urgency builds up in the agent, eventually compelling it to commit to one of the choice alternatives even when evidence is not particularly great. This urgency signal has been implemented as an additive term that added to the accumulated evidence. This urgency term monotonically increases with time does not depend on stimulus uncertainty and has been supported by recent neurobiological evidence \cite{thura_basal_2017, cisek_decisions_2009, hanks_neural_2014}. 

Inspired by the idea of the urgency signal \cite{cisek_decisions_2009}, we modified our basic model by adding an urgency term to the stimulus evidence, replacing $E_{t}$ in Eq.  \eqref{eqn:state}  by $U_{t}$ defined such that:

\begin{equation}  
\label{eqn:urgency}
      U_{t} = E_{t} + {\rm sgn}(E_{t}) \rho t
\end{equation}

where $E_{t}$ = $\mathcal{N}(Kc^u,\,\sigma^{2})$, $\rho$ is the normalization factor (set to 0.005), ${\rm sgn}$ is the sign function and $t$ is time within a trial \cite{cisek_decisions_2009}. In this formulation, the second term is a time-dependant signal that is added to the stimulus evidence ($E_{t}$). Replacing Eq.  \eqref{eqn:state} with Eq.  \eqref{eqn:urgency}, we obtained a new model in which the state variable accumulates both time and evidence. Our simulations showed that this urgency model does reproduce the predicted longer error RT (Fig. \ref{fig:err_corr}(b)). Importantly, comparing the position of terminal states in the Q-tables in correct and error trials we observe that terminal states are similar (Fig.  \ref{fig:err_corr}(b) inset). Given that the position of the terminal states corresponds to what DDM literature calls the decision boundary, our simulation results are in line with the previous physiological findings \cite{purcell_neural_2016, okazawa_neural_2023, roitman_response_2002} that reported  identical level of neural firing for boundary neurons in correct and error trials.  
 
\subsubsection{Post Error Slowing}
Post Error slowing (PES) is one of the oldest and most well-known behavioural observations in decision sciences \cite{logan_cognitive_2010}.  After an error, reaction time in the next trial is slower compared to a trial of the same level of difficulty that follows from a correct outcome. This intuitively understandable and well-known behavioural effect has presented substantial challenges to the computational and neurobiological studies of decision making. 

In one study, Purcell et al \cite{purcell_neural_2016}, investigated PES using the same RDK paradigm that we have focused on here and reported that although post-error reaction times were longer than post-correct, accuracy was not different between them. They also examined the monkey LIP's firing rates at the time of response in order to compare the decision threshold in post-error and post-correct trials and did not observe any difference between them. These two sets of findings are problematic because, within the framework of sequential sampling models including DDM, PES is clearly explained as an adjustment of the decision threshold in the current trial based on the previous outcome. This prediction entails that accuracy should also be higher after an error vs after a correct choice. Purcell et al's empirical observations do not fit this prediction, neither at the level of behavior nor neural activity. An alternative suggestion that has been proposed to explain PES is that following an error, sensitivity (i.e., drift rates in DDM) could be modified. However, the empirical observations do not fit this idea either. To account for their findings, Purcell et al. proposed a model in which changes in \textit{urgency} (implemented as collapsing bound) were combined with changes in sensitivity to obtain a model that delivers a change in reaction time but preserves accuracy in post-error trials (see figure 2 of their study \cite{purcell_neural_2016}). 

Earlier, in  \ref{sec:urgency} we introduced urgency in our framework. Here we show that the same formulation, when combined with the outcome from the previous trial provides a more parsimonious account that is consistent with \cite{purcell_neural_2016} without having to assume any changes in sensitivity.  Following \cite{purcell_neural_2016}, we define the level of urgency (Eq.  \eqref{eqn:urgency}, also see \ref{sec:urgency}) as a function of outcome in the previous trial as follows: 

\begin{equation}  
\label{eqn:PES}
    \rho_{u} = 
    \begin{cases}
        0.001 & \text{if } R_{u-1} < -1 \\
        0.003 & \text{if } R_{u-1} >  0
    \end{cases}
\end{equation}
Where $R_{u-1}$ is the reward in trial $u-1$ and $\rho_{t}$ is the normalization factor, in trial $u$ in equation \eqref{eqn:urgency}. 
 
\begin{figure}
\centering
\includegraphics[width=.99\textwidth]{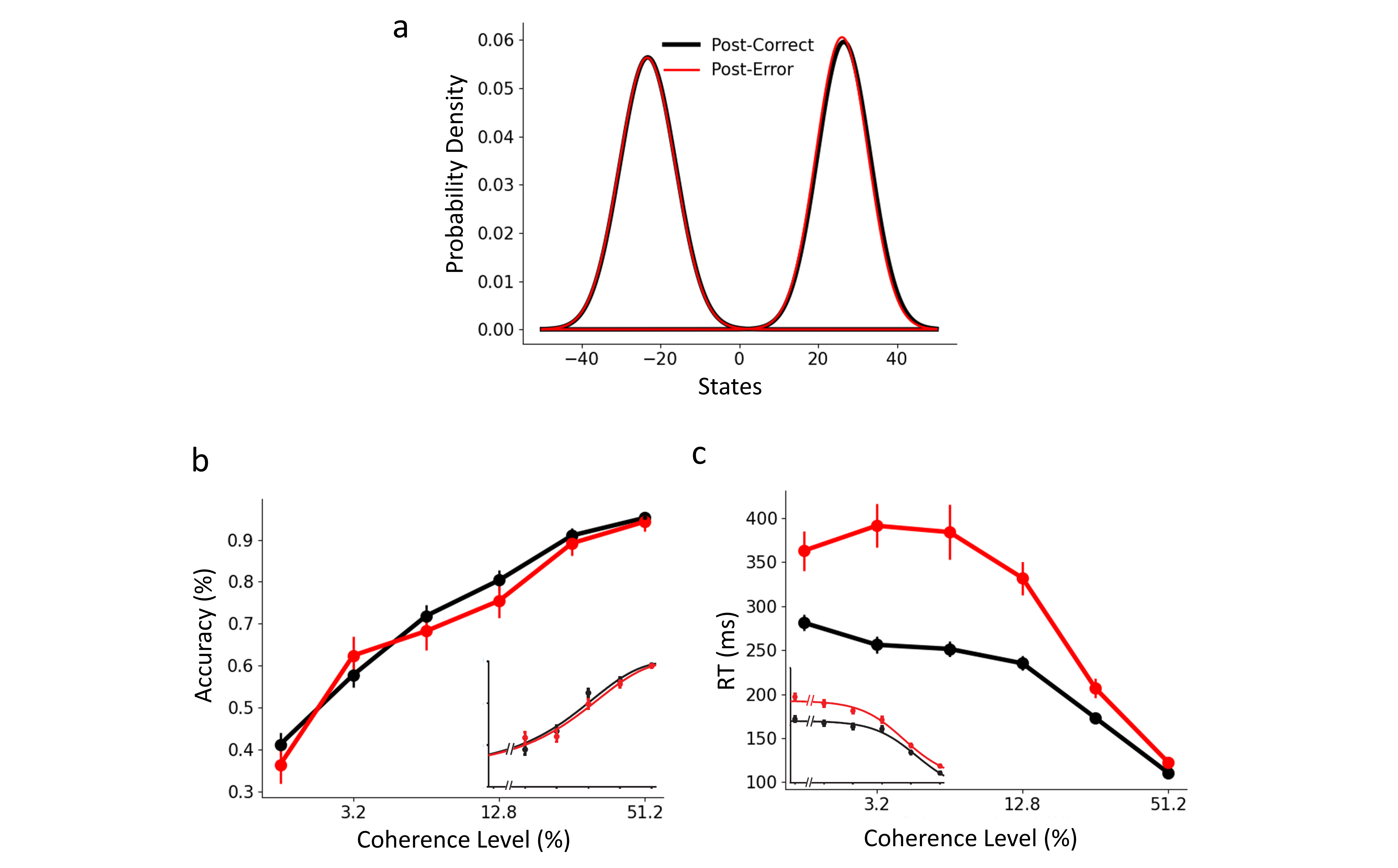}
\caption{\label{fig:PES}\textbf{Replication the Post Error Slowing.} (a) Distribution of terminal states in post-error (red) vs post-correct (black) trials. (b) Accuracy of post error vs post correct trials. The accuracy did not change, similar to a previous study. (c) Same as (b) but for RT. Post error trials are significantly slower than the post correct trials; in line with what has been reported before (inset). Error bars are SEM across simulations (U = 2400). Insets show empirical data from \cite{purcell_neural_2016}}
\end{figure}

The results of the model simulation (number of trials $U = 2400$; learning rate = 0.1;  reward set \{${20, -50, -1}$\} for correct, error and wait, respectively) are plotted in  (Fig.  \ref{fig:PES}). Data from the second part of the simulation (i.e., trial 1200 to 3600) is shown here. Each trial has been labeled according to its preceding outcome. Several observations are evident. Fig. \ref{fig:PES}(a) shows that the terminal states (i.e., decision threshold) do not differ between the post-error and post-correct trials. As would be expected from this fact, at the level of behaviour, no difference is observed in accuracy (Fig. \ref{fig:PES}(b)). RTs, however, are longer in post-error trials (Fig. \ref{fig:PES}(c)).  These findings are in line with previous empirical observations e.g., ref. \cite{purcell_neural_2016} . The key difference here is that the model proposed here is simpler and more parsimonious involving outcome-dependent modulation of urgency without requiring any modulation of sensitivity. 

\subsubsection{Impact of evidence volatility on behavior}
\label{vol_rep}
\begin{figure}
\centering
\includegraphics[width=.99\textwidth]{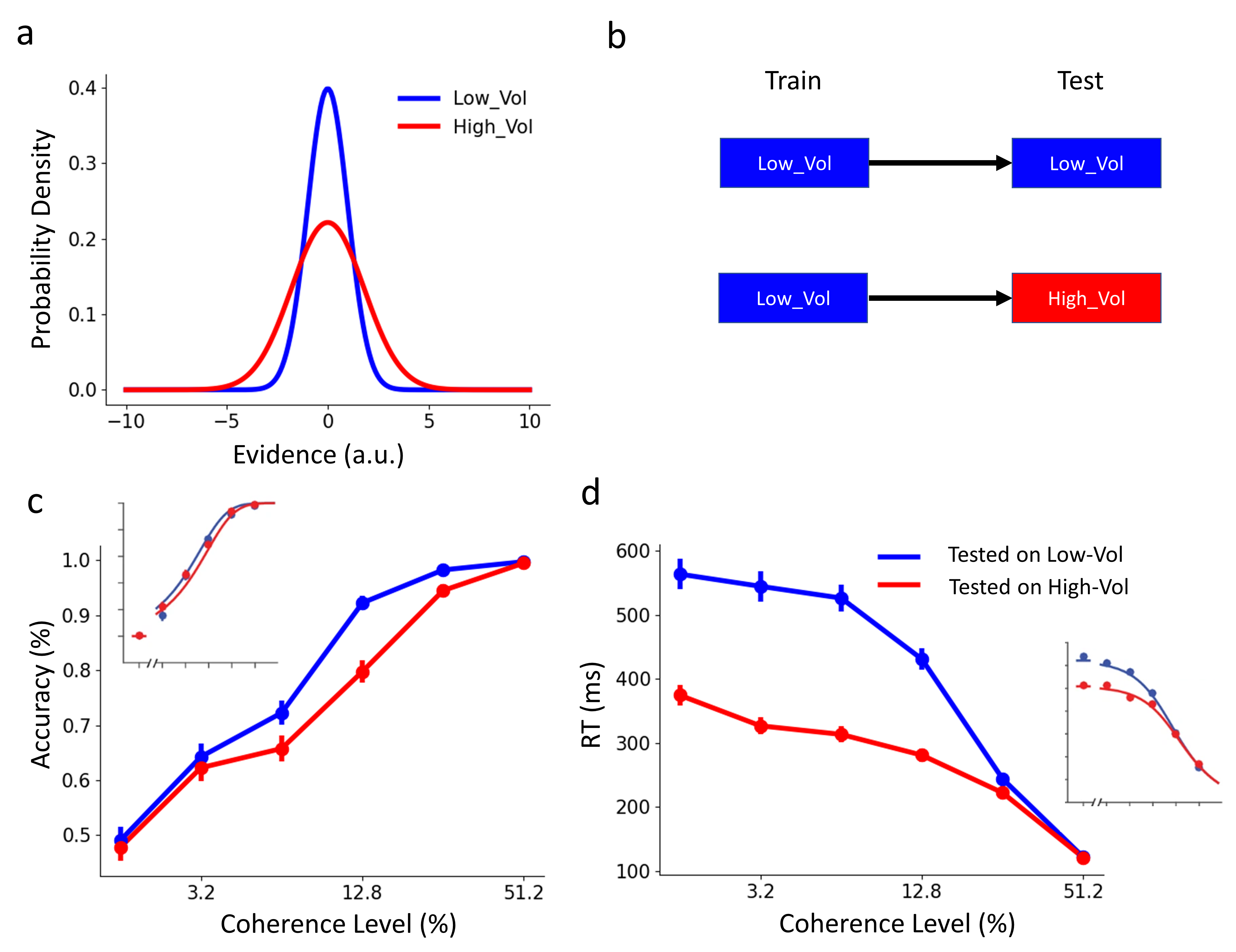}
\caption{\label{fig:vol_replicate} \textbf{Replication of the volatility effect.} (a) Depiction of the distribution of evidence in two volatility conditions in 0\% coherence level. Both conditions have a similar mean (zero) but the variance (volatility) is higher in the high-volatility condition (red) than in the low-volatility condition (blue). (b) The procedure of our simulations. In both volatility conditions, we first trained the model on low-volatility evidence. Then we test the model based on high (red) and low (blue) volatility trials.  (c-d) Model behaviors in different volatility conditions. Test accuracy (c) and RT (d) of the model are lower in high volatility condition (red) similar to previous findings \cite{zylberberg_influence_nodate} (insets). Error bars are SEM across trials (U =  2400).}
\end{figure}

Studies examining perceptual decision making under uncertainty most often look at the relationship between the first moment, i.e., \textit{mean} of the perceptual evidence (e.g., coherence level in random dot motions, luminance contrast in oriented gratings, vibration amplitude in somatosensory psychophysics) and the behavior \cite{shadlen_decision_2013, gold_neural_2007, britten_analysis_1992}. This relationship is indeed captured by our basic model too (\ref{fig:trained_model}(a)-(b)). Fewer studies have examined the role of the second moment i.e., variance  (but see here for a review of second-order perception \cite{baker_chapter_2001}).  A recent study   \cite{zylberberg_influence_nodate} examined human behavior (accuracy and reaction times) when the mean perceptual evidence was kept constant but its volatility was systematically manipulated. Under high volatility, both accuracy and RT were, reduced (Fig. \ref{fig:vol_replicate}.c-d inset).

Following \cite{zylberberg_influence_nodate}, we first created two volatility conditions (Fig. \ref{fig:vol_replicate}(a), low volatility (blue): $\sigma = 1$ in $E_{t}$ in Eq.  \eqref{eqn:state}, high volatility (red): $\sigma = 1.3$). Then, we trained our basic model on low volatility conditions (Fig. \ref{fig:vol_replicate}(b)) (learning rate = 0.1, \textit{U}=2400 and tested the trained model under two different conditions of volatility (Fig. \ref{fig:vol_replicate}(b)). Our simulations showed that under high volatility testing conditions, reaction times and accuracy both decreased (Fig. \ref{fig:vol_replicate}(c)-(d)) thereby neatly replicating the empirical findings of \cite{zylberberg_influence_nodate}. 

\end{document}